\documentclass{article}

\PassOptionsToPackage{numbers, compress}{natbib}

\usepackage[final]{neurips_2025}

\usepackage[utf8]{inputenc} 
\usepackage[T1]{fontenc}    
\usepackage{hyperref}       
\usepackage{url}            
\usepackage{booktabs}       
\usepackage{amsfonts}       
\usepackage{nicefrac}       
\usepackage{microtype}      
\usepackage{xcolor}         

\usepackage{algorithm}
\usepackage{algpseudocode}

\usepackage{color, colortbl}
\usepackage{graphicx}
\usepackage{amsmath}
\usepackage{multicol}
\usepackage{multirow}
\usepackage{arydshln}
\usepackage[capitalize]{cleveref}
\usepackage{amssymb}
\usepackage{amsthm}
\usepackage{booktabs}
\usepackage{pifont}
\usepackage{subcaption}
\usepackage{wrapfig}
\newcommand{\cmark}{{\ding{51}}}
\newcommand{\xmark}{{\ding{55}}}
\definecolor{Gray}{gray}{0.9}

\newcolumntype{g}{>{\columncolor{Gray!80}}c}

\newtheorem{theorem}{Theorem}
\newtheorem{prop}{Proposition}

\newcommand{\dashmidrule}{
  \hdashline
  \addlinespace[1pt]
}

\def\framework{APDM}

\def\methodTwo{L2P}
\def\eg{\emph{e.g. }}
\def\ie{\emph{i.e. }}

\title{Perturb a Model, Not an Image: \\ Towards Robust Privacy Protection \\ via Anti-Personalized Diffusion Models}

\author{%
  Tae-Young Lee$^1$\thanks{Equal contribution.}
  \quad
  Juwon Seo$^{2}$\footnotemark[1]
  \quad
  Jong Hwan Ko$^3$\thanks{Corresponding authors.}
  \quad
  Gyeong-Moon Park$^1$\footnotemark[2]
  \\
  \\
  $^1$Korea University 
  \quad
  $^2$Kyung Hee University 
  \quad
  $^3$Sungkyunkwan University \\ \addlinespace[2pt]
  \texttt{tylee0415@korea.ac.kr} 
  \quad
  \texttt{jwseo001@khu.ac.kr} \\
  \texttt{jhko@skku.edu}
  \quad
  \texttt{gm-park@korea.ac.kr}
}

\begin{document}

\maketitle
\begin{abstract}
Recent advances in diffusion models have enabled high-quality synthesis of specific subjects, such as identities or objects.
This capability, while unlocking new possibilities in content creation, also introduces significant privacy risks, as personalization techniques can be misused by malicious users to generate unauthorized content. 
Although several studies have attempted to counter this by generating adversarially perturbed samples designed to disrupt personalization, they rely on unrealistic assumptions and become ineffective in the presence of even a few clean images or under simple image transformations.
To address these challenges, we shift the protection target from the images to the diffusion model itself to hinder the personalization of specific subjects, through our novel framework called \textbf{A}nti-\textbf{P}ersonalized \textbf{D}iffusion \textbf{M}odels (\textbf{APDM}). 
We first provide a theoretical analysis demonstrating that a naive approach of existing loss functions to diffusion models is inherently incapable of ensuring convergence for robust anti-personalization.
Motivated by this finding, we introduce Direct Protective Optimization (DPO), a novel loss function that effectively disrupts subject personalization in the target model without compromising generative quality.
Moreover, we propose a new dual-path optimization strategy, coined Learning to Protect (L2P). 
By alternating between personalization and protection paths, L2P simulates future personalization trajectories and adaptively reinforces protection at each step.
Experimental results demonstrate that our framework outperforms existing methods, achieving state-of-the-art performance in preventing unauthorized personalization.
The code is available at \href{https://github.com/KU-VGI/APDM}{https://github.com/KU-VGI/APDM}.
\end{abstract}

\vspace{-3mm}
\section{Introduction}
\label{sec:intro}
\vspace{-1mm}

\begin{figure}[!t]
    \centering
    \includegraphics[width=\linewidth]{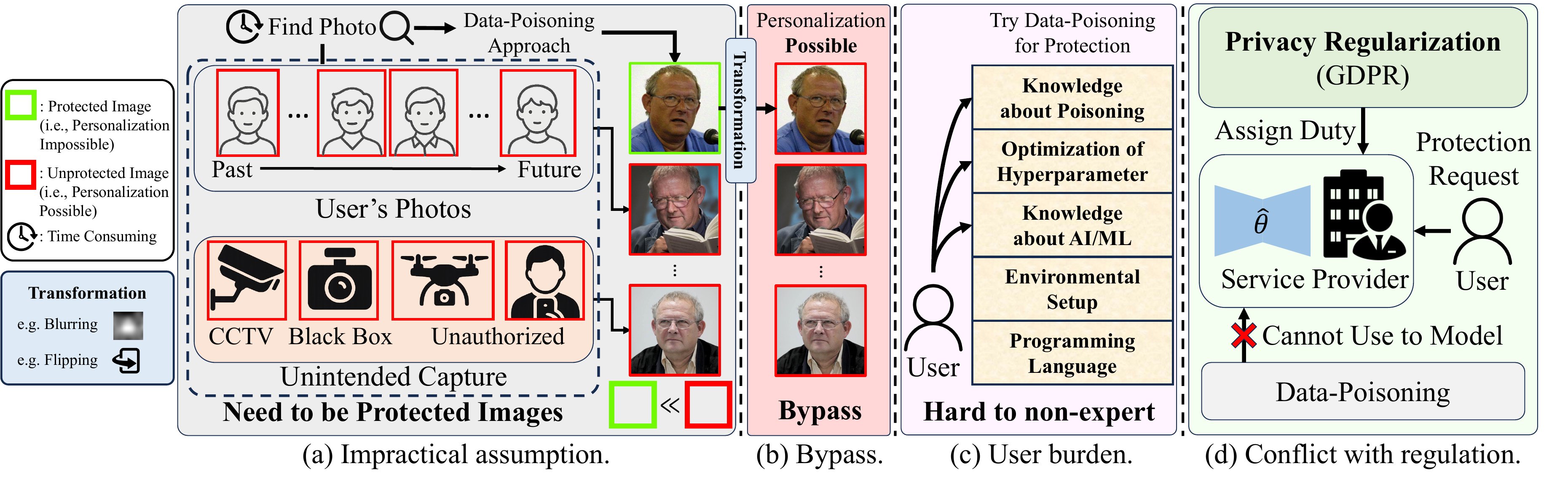}
    \vspace{-6mm}
    \caption{\textbf{Motivation Figure.} Existing protection approaches face critical limitations: (a) \textit{impracticality} of applying data-poisoning to all images, (b) \textit{vulnerability to easy circumvention} of protection methods, (c) \textit{high entry barriers} for non-expert users, and (d) \textit{incompatibility with service providers} who must comply with privacy regulations.}
    \vspace{-6mm}
    \label{fig:motivate}
\end{figure}

Diffusion models (DM)~\cite{sohl2015deep, ho2020denoising} have become prominent generative models across various domains and tasks, including image, video, and audio synthesis~\cite{rombach2022high, guo2024animatediff, liu2023audioldm}, image-to-image translation~\cite{parmar2023zero}, and image editing~\cite{hertz2023prompt}.
Among these, personalization techniques~\cite{gal2023an, ruiz2023dreambooth, kumari2023multi}—enabling the generation of images depicting specific subjects (\eg individuals, objects) in varied contexts, such as \textit{``an image of my dog on the moon''}—have received significant attention.
Several approaches, such as DreamBooth~\cite{ruiz2023dreambooth} and Custom Diffusion~\cite{kumari2023multi}, have demonstrated highly effective capabilities for personalized image generation.
However, such personalization also presents substantial privacy risks, as malicious users could exploit it to create unauthorized images of specific individuals, for instance, to generate and distribute fake news, thereby raising significant social and ethical concerns.

To prevent misuse of such personalization capability from a user's request, several protection approaches~\cite{liang2023adversarial, van2023anti, Wang2024simac, wan2024prompt} based on data-poisoning have been proposed.
They directly add imperceptible noise perturbations to the images of the specific subject using the Projected Gradient Descent (PGD)~\cite{madry2018towards}.
When a malicious user attempts to personalize using these perturbed images, the added noise disrupts the stability of the training process, resulting in ineffective personalization convergence.

However, existing approaches suffer from several critical limitations in real-world scenarios (Figure~\ref{fig:motivate}). 
Most importantly, their efficacy often hinges on the \textit{impractical assumption} that users can apply poisoning comprehensively across their personal image collections—including those already shared, newly created, or even unintentionally captured—which is a practically unachievable task.
This limitation enables malicious users to \textit{easily bypass protection} using unprotected images.
Furthermore, even if the images are perturbed, attackers can still circumvent defense by applying transformations that weaken the perturbation effects~\cite{van2023anti, liu2024metacloak, honig2025adversarial}.
On the other hand, data-poisoning is predominantly a user-centric defense, placing the \textit{implementation burden on individuals} who are often non-experts, making widespread adoption unrealistic. 
Furthermore, this user-level design of existing approaches \textit{conflicts with privacy regulations}-such as the GDPR~\cite{voigt2017gdpr}-that assign service providers the obligation to ensure anti-personalization upon user requests.
As a result, such methods are inherently unsuitable for provider-side deployment (see~\cref{supp:add_motivate} for more details).

\vspace{-1mm}
Taken together, these issues highlight the need to move beyond user-side defenses toward model-level solutions that not only enable service providers to enforce anti-personalization directly within their systems but also enhance robustness and practicality in real-world deployments.
To address this, we shift our focus from the data samples to the DMs themselves. 
In this paper, we propose \textbf{A}nti-\textbf{P}ersonalized \textbf{D}iffusion \textbf{M}odel (\textbf{\framework{}}), a novel framework designed to directly remove personalization capabilities for specific subjects within pre-trained DMs, without data-poisoning. 
The primary goals of \framework{} are twofold: (i) \textit{preventing} the unauthorized personalization attempts, resulting in failed or irrelevant generations, and (ii) \textit{preserving} the generation performance and its ability to personalize other, non-targeted subjects. 
To the best of our knowledge, \framework{} is the first approach to directly update the model parameters for protection, inherently overcoming data dependency.


\vspace{-1mm}
However, simply redirecting the protection effort to the model parameters does not guarantee success if we na\"ively adopt strategies from data-centric methods. 
Firstly, we theoretically prove that directly applying loss—originally designed for creating adversarial perturbations on images—to the model parameters fails to converge. 
To this end, we introduce a novel loss function, \textbf{D}irect \textbf{P}rotective \textbf{O}ptimization (\textbf{DPO}), disrupting the personalization process.
Moreover, simply applying a protection loss uniformly is insufficient, since personalization involves iterative updates to model parameters. 
Therefore, being aware of the personalization trajectory is essential for robust protection.
For this reason, we propose \textbf{L}earning to \textbf{P}rotect (\textbf{L2P}), a dual-path optimization strategy. 
L2P alternates between a personalization path, simulating potential future personalized model states, and a protection path, which leverages these intermediate states to apply adaptive, trajectory-aware protective updates. 
This dynamic approach allows the model to anticipate and counteract personalization attempts, ensuring robust DM protection in across various scenarios.

\vspace{-1mm}
Our contributions can be summarized as follows:
\vspace{-2mm}
\begin{itemize}
    \item For the first time, we propose a novel framework, called \textbf{A}nti-\textbf{P}ersonalized \textbf{D}iffusion \textbf{M}odel (\textbf{\framework{}}), for robust anti-personalization in DMs by directly updating \textit{model parameters}, unlike existing data-centric methods. This approach fundamentally overcomes the impractical assumptions and data dependency issues of prior works.
    \item We theoretically prove that a naive application of existing image perturbation losses directly to model parameters fails to converge. To address this, we propose a novel objective, \textbf{D}irect \textbf{P}rotective \textbf{O}ptimization (\textbf{DPO}) loss. DPO guides the model to remove the personalization capability of a specific subject while preserving generation performance.
    \item To effectively counteract the iterative and adaptive process of personalization, we introduce \textbf{L}earning to \textbf{P}rotect (\textbf{L2P}), a dual-path optimization strategy that anticipates personalization trajectories and reinforces protection accordingly, enabling robust defense.
    \item We empirically demonstrate that \framework{} can safeguard against personalization in real-world scenarios, achieving state-of-the-art performance across various personalization subjects.
\end{itemize}

\vspace{-4mm}
\section{Related Work}
\vspace{-2mm}
\paragraph{Personalized Text-to-Image Diffusion Models.}
The advancement of diffusion-based image synthesis, like Stable Diffusion (SD)~\cite{rombach2022high}, has enabled not only high-quality image generation but also the creation that reflect desired contexts from the text.
This advancement has accelerated the widespread application of Text-to-Image (T2I) DMs~\cite{rombach2022high}, one of which is personalization, such as generating images containing specific objects under the various situations (\eg a particular dog or person on the moon).
Consequently, research on personalized models has emerged.
The most widely used method is DreamBooth~\cite{ruiz2023dreambooth}, which fine-tunes a pre-trained SD using a small set of images depicting a specific concept (\eg a particular person).
This allows users to generate desired images containing the target object.
Texture Inversion~\cite{gal2023an} achieves this by searching for an optimal text embedding that can represent the target object based on pseudo-words.
Custom Diffusion~\cite{kumari2023multi} optimizes the key and value projection matrices in the cross-attention layers of the pre-trained SD, offering more efficient and robust personalization performance.
However, these methods are a double-edged sword, offering powerful personalization but also posing risks, such as misuse in crimes or unintended applications.
\vspace{-0.3cm}
\paragraph{Protection against Unauthorized Personalization.}
To prevent unauthorized usage, many protection methods have been developed based on adversarial attacks \cite{goodfellow2015explaining, carlini2017towards, madry2018towards}.
AdvDM~\cite{liang2023adversarial} was the first to extend classification-based adversarial attack methods to DMs, generating adversarial samples for protecting personalization.
Furthermore, Anti-DreamBooth~\cite{van2023anti} proposed protection against more challenging fine-tuned DMs (\eg DreamBooth).
They used a fine-tuned surrogate model as guidance to obtain optimal perturbations for adversarial images.
SimAC~\cite{Wang2024simac} improved this optimization process to better suit DMs, while CAAT~\cite{xu2024perturbing} focused on reducing time costs by updating cross-attention blocks.
MetaCloak~\cite{liu2024metacloak} and PID~\cite{li2024pid} have also been conducted to counter text variation or image transformation techniques (\eg filtering).
The most recent work, PAP~\cite{wan2024prompt}, tries to predict potential prompt variations using Laplace approximation.
However, existing works have primarily focused on how to effectively add perturbations to images for protection.
In contrast, as we mentioned above, we apply protection directly at the model level, reflecting real-world demands.
\section{Preliminaries}
\vspace{-2mm}
\subsection{Text-to-Image Diffusion Models}
\vspace{-2mm}
T2I DMs \cite{rombach2022high}, a popular variant of DMs~\cite{ho2020denoising, sohl2015deep} generate an image $\hat{x}_0$ corresponding to a given text prompt embedding $c$.
T2I DMs operate via forward and reverse processes.
In the forward process, noise $\epsilon\sim\mathcal{N}(0,I)$ is added to input image $x_0$ to produce noisy image $x_t$ at a timestep $t\in[0,T]$:
\vspace{-1mm}
\begin{align}
\label{eq:add_noise}
    x_t = \sqrt{\bar{\alpha}_t}x_0 + \sqrt{1 - \bar{\alpha}_t}\epsilon,
\end{align}
where $\bar\alpha_t=\Pi^t_{i=1}\alpha_i$ is computed from noise schedule $\{\alpha_t\}^T_{t=0}$.
In the reverse process, DM, parameterized by $\theta$, aims to denoise $x_t$.
DM is trained to predict the noise residuals added to $x_t$:
\begin{align}
\label{eq:l_simple}
\mathcal{L}_{simple}=\mathbb{E}_{x_0,t,c,\epsilon\sim\mathcal{N}(0,I)}{\Vert \epsilon_\theta(x_t,t,c) - \epsilon\Vert}^2_2.
\end{align}

\vspace{-4mm}
\subsection{Personalized Diffusion Models}
\vspace{-1mm}
To generate images that include a specific subject, several works personalize pre-trained T2I DMs \cite{ruiz2023dreambooth, kumari2023multi}.
Given a small image set $x_0\in\mathcal{X}$ of the subject and a text embedding $c^{per}$ with a unique identifier, \eg \textit{``a photo of [V*] person''}, they modify the loss function in Eq.\eqref{eq:l_simple} as follows:
\begin{align}
\label{eq:l_db}
\mathcal{L}^{per}_{simple}=\mathbb{E}_{x_0,t,c^{per},\epsilon\sim\mathcal{N}(0,I)}{\Vert \epsilon_\theta(x_t,t,c^{per}) - \epsilon\Vert}^2_2,
\end{align}
where $x_t$ is a noisy image from Eq.~\eqref{eq:add_noise}.
However, directly applying this modified loss can cause language drift, where the personalized DM generates images related to target subject, even without unique identifier.
To mitigate this, DreamBooth~\cite{ruiz2023dreambooth} introduces a prior preservation loss function that leverages the pre-trained DM.
This encourages DM, using a class-specific text embedding $c^{pr}$ (\eg \textit{``a photo of person''}), to retain its knowledge of the general class associated with the specific subject:
\vspace{-1mm}
\begin{align}
    \label{eq:l_ppl}
    \mathcal{L}_{ppl}=\mathbb{E}_{x^{pr}_0,t,c^{pr},\epsilon\sim\mathcal{N}(0,I)}{\Vert \epsilon_\theta(x^{pr}_t,t,c^{pr}) - \epsilon\Vert}^2_2,
\end{align}
where $x^{pr}_0$ is a generated sample from the pre-trained T2I DM with the text embedding $c^{pr}$, and $x^{pr}_t$ is the noisy version of $x^{pr}_0$ at timestep $t$.
Alternatively, Custom Diffusion \cite{kumari2023multi} utilizes images from training dataset instead of generated images for $x^{pr}$.
The final objective for personalization becomes:
\vspace{-3mm}
\begin{align}
\mathcal{L}_{per}=\mathcal{L}^{per}_{simple} + \mathcal{L}_{ppl}.
\label{eq:l_per}
\end{align}
\vspace{-8mm}
\vspace{-1mm}
\section{Method}
\vspace{-1mm}
\subsection{Problem Formulation}
\vspace{-1mm}
\label{sec:4-1}
Unlike prior approaches that perturb \textit{images}, we directly update the \textit{parameters $\theta$} of the pre-trained DM using only a small image set $x_0 \in \mathcal{X}$.
Our goal is to transform $\theta$ into a safeguarded model $\hat\theta$ that inherently resists personalization of the subject appearing in these images.
This process can be viewed as optimizing the model parameters with respect to a protection objective:
\vspace{-1mm}
\begin{align}
    \hat\theta = \arg\min_\theta\mathcal{L}_{protect},
\end{align}
\vspace{-1mm}
where $\mathcal{L}_{protect}$ is a loss function to prevent personalization, which will be discussed in Section \ref{sec:4-2-1}.
Subsequently, if an adversary attempts to personalize a subject in $\mathcal{X}$ with this safeguarded model $\hat\theta$, the resulting personalized model $\hat\theta_{per}$ is obtained as follows: 
\vspace{-1mm}
\begin{align}
    \hat\theta_{per} = \arg\min_{\hat\theta}\mathcal{L}_{per}.
\end{align}
Our approach has two main objectives.
For protection, the re-personalized model $\hat\theta_{per}$ should yield low-quality images or images of subjects perceptually distinct from those in $\mathcal{X}$.
For stability, the protected model $\hat\theta$ should be able to generate high-quality images and effectively personalize for the other subjects, comparable to those produced by the pre-trained DM $\theta$.

\vspace{-1mm}
\subsection{Analysis of Na\"ive Approach}
\vspace{-1mm}
\label{sec:4-2}
A naive yet intuitive way to protect the model is to extend existing data-poisoning approaches~\cite{liang2023adversarial,van2023anti,Wang2024simac,wan2024prompt} to the model level.
Specifically, their noise update process that maximizes $\mathcal{L}^{per}_{simple}$ using PGD~\cite{madry2018towards} can be naturally applied at the model level.
In addition, the model’s generative performance can be preserved by incorporating $\mathcal{L}_{ppl}$, as done in DreamBooth~\cite{ruiz2023dreambooth}.  
The overall objective for this na\"ive approach can be expressed as follows:
\begin{align} 
\label{eq:adv} 
    \mathcal{L}_{adv}=-\mathcal{L}^{per}_{simple} + \mathcal{L}_{ppl}.
\end{align}
To ensure effective protection using $\mathcal{L}_{adv}$, the optimization process must converge.
We analyze the necessary conditions for convergence by examining the gradients of the loss with respect to $\theta$.
This leads to the following Proposition~\ref{prop:convergence condition} (proof in~\cref{supp:proof_prop_1}).
\begin{prop}
A necessary condition for $\mathcal{L}_{adv}$ to converge to a local minimum with respect to model parameters $\theta$ is that the gradients of its constituent terms, $\nabla_\theta \mathcal{L}_{simple}^{per}$ and $\nabla_\theta \mathcal{L}_{ppl}$, must point in the same direction.
\label{prop:convergence condition}
\end{prop}
To further understand how these gradients influence each other during optimization, we analyze their interaction through the first-order Taylor approximation and derive the following relationships.
\begin{align}
    \label{eq:preservation_relation}
    (\nabla_\theta\mathcal{L}_{simple}^{per}(\theta))^\top\cdot 
    (\nabla_\theta\mathcal{L}_{ppl}(\theta))) 
    < \|\nabla_\theta\mathcal{L}_{ppl}(\theta))\|^2, \\
    \label{eq:personalization_relation}
    (\nabla_\theta\mathcal{L}_{simple}^{per}(\theta))^\top\cdot 
    (\nabla_\theta\mathcal{L}_{ppl}(\theta))) 
    < \|\nabla_\theta\mathcal{L}_{simple}^{per}(\theta)\|^2.
\end{align}
Based on the Proposition~\ref{prop:convergence condition}, we can restrict the left terms in Eq.~\eqref{eq:preservation_relation} and~\eqref{eq:personalization_relation}, as $|\nabla_\theta\mathcal{L}_{simple}^{per}(\theta)|\cdot|\nabla_\theta\mathcal{L}_{ppl}(\theta)|$. Using these results, we can rewrite the Eq.~\eqref{eq:preservation_relation} and~\eqref{eq:personalization_relation} as:
\begin{align}
    \label{eq:preservation_inequality}
    |\nabla_\theta\mathcal{L}_{simple}^{per}(\theta)|
    < |\nabla_\theta\mathcal{L}_{ppl}(\theta))|, \\
    \label{eq:personalization_inequality}
    |\nabla_\theta\mathcal{L}_{ppl}(\theta)| 
    < |\nabla_\theta\mathcal{L}_{simple}^{per}(\theta)|.
\end{align}

By combining Proposition~\ref{prop:convergence condition} with the inequalities above, we observe that the required gradient alignment for convergence cannot hold, which we formalize in the following theorem (see~\cref{supp:proof_theo_1}).
\begin{theorem}
    If the objective is to simultaneously reduce both $-\mathcal{L}_{simple}^{per}$ and $\mathcal{L}_{ppl}$, the necessary condition for convergence outlined in Proposition~\ref{prop:convergence condition} leads to the contradictory requirements presented in Eq.\eqref{eq:preservation_inequality} and \eqref{eq:personalization_inequality}.
    Therefore, $\mathcal{L}_{adv}$ composed of such conflicting terms generally fails to converge to a point that effectively optimizes both objectives.
\label{theo:theorem1}
\end{theorem}
Therefore, a new loss function is required to resolve this conflict and ensure that anti-personalization updates stay consistent with the denoising process, maintaining both generation quality and protection.


\begin{figure}[t]
    \centering
    \includegraphics[width=\textwidth]{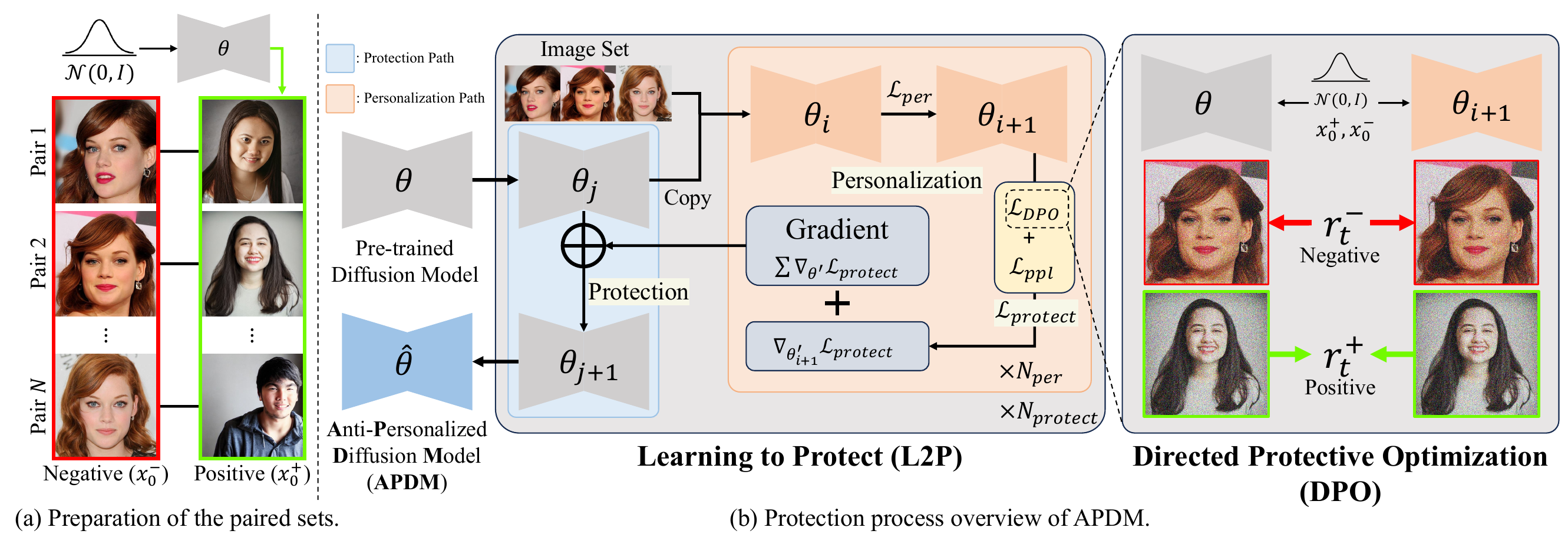}
    \vspace{-6mm}
    \caption{\textbf{Overview}. To prevent personalization in the parameter level, we propose Anti-Personalized Diffusion Model (APDM). (a) APDM first generates a paired image for each clean input image $x_0$. (b) APDM consists of two components - (i) Learning to Protect, a novel optimization algorithm that makes the protection procedure aware of personalization trajectories, and (ii) Directed Protective Optimization loss, designed to disrupt personalization while preserving the generation capabilities.}
    \label{fig:l2p}
    \vspace{-2mm}
\end{figure}
\subsection{Anti-Personalized Diffusion Models}
\label{sec:4-3}
To achieve the dual goals outlined in \ref{sec:4-1}, we propose a novel framework, \textbf{A}nti-\textbf{P}ersonalized \textbf{D}iffusion \textbf{M}odels (\textbf{APDM}).
\framework{} introduces a novel loss function, called Direct Protective Optimization (DPO), which aims to prevent personalization in DMs while maintaining their original generative performance (Section \ref{sec:4-2-1}).
DPO effectively mitigates the model collapse issue discussed in Section~\ref{sec:4-2}.
Furthermore, we propose a novel dual-path optimization scheme, Learning to Protect (\methodTwo{}), which considers the trajectory of personalization during training to apply the proposed loss function more effectively (Section \ref{sec:4-3-2}).
The overview of APDM is presented in Figure~\ref{fig:l2p}.

\subsubsection{Direct Protective Optimization}
\vspace{-1mm}
\label{sec:4-2-1}
Instead of $\mathcal{L}_{adv}$, which degrades the model's distribution due to convergence failure (Section~\ref{sec:4-2}), we directly guide the model on which information should be learned and which should be suppressed.
Inspired by Direct Preference Optimization~\cite{rafailov2023direct}, given a pair of images $(x^+_0, x^-_0)$, we designate $x^+_0$ as a positive sample to be encouraged during the protection procedure and $x^-_0$ as a negative sample to be discouraged, \ie an image containing a specific subject to be protected ($x_0 \in \mathcal{X}$).
By incorporating the Bradley-Terry model, the probability of preferring $x^+_0$ over $x^-_0$ can be expressed as: 
\begin{align} 
\label{eq:eq13}
    p(x^+_0 > x^-_0)=\sigma(r(x^+_0) - r(x^-_0)),
\end{align}
where $\sigma(\cdot)$ denotes the sigmoid function and $r(\cdot)$ represents the reward function.
Building upon the formulation of Diffusion-DPO~\cite{wallace2024diffusion} (see~\cref{supp:derivation} for detailed derivation), we define a new \textbf{Direct Protective Optimization} (\textbf{DPO}) as follows: 
\vspace{-1mm}
\begin{align}
\label{eq:eq14}
    \begin{split}
        &r^+=\Vert\epsilon_\theta(x^+_t,t,c)-\epsilon\Vert^2_2-\Vert\epsilon_\phi(x^+_t,t,c)-\epsilon\Vert^2_2, \\
        &r^-=\Vert\epsilon_\theta(x^-_t,t,c)-\epsilon\Vert^2_2-\Vert\epsilon_\phi(x^-_t,t,c)-\epsilon\Vert^2_2, \\
        &\mathcal{L}_{DPO}=-\mathbb{E}_{x^+_0,x^-_0,c,t,\epsilon \sim N(0,I)} \log \sigma(-\beta(r^+-r^-)),
    \end{split}
\end{align}
where $\phi$ is a pre-trained DM and $\beta$ is a hyper-parameter that controls the extent to which $\theta$ can diverge from $\phi$.
In our DPO, we prepare $x^+_0$ by synthesizing images from pre-trained T2I DMs $\phi$ using a generic prompt $c^{pr}$, and they are paired one-to-one with the negative samples $\mathcal{X}$.
This approach naturally encourages the generation of generic (positive) images while effectively suppressing the synthesis of negative images depicting the specific subject.

Finally, combining the proposed loss term with the preservation loss ($\mathcal{L}_{ppl}$), the final objective is:
\vspace{-1mm}
\begin{align}
    \mathcal{L}_{protect}=\mathcal{L}_{DPO}+\mathcal{L}_{ppl}.
\end{align}

\subsubsection{Learning to Protect}
\label{sec:4-3-2}
Since the personalization of DMs involves iterative updates to model parameters, effective protection should consider the evolving personalized states at different states.
Therefore, instead of simply applying our $\mathcal{L}_{protect}$ uniformly to the model, we simulate the future personalization path in advance, allowing the model to anticipate upcoming parameter changes during personalization.
To this end, we introduce a novel dual-path optimization algorithm, \textbf{Learning to Protect} (\textbf{\methodTwo{}}).
\methodTwo{} integrates personalization into the protection loop, enabling the model to learn from simulated personalization behaviors and adjust its parameters for adaptive and robust protection.

\begin{algorithm}[t]
\caption{Learning to Protect (L2P)}
\label{algorithm}
\textbf{Input:} pre-trained model $\theta$, loss function for personalization $\mathcal{L}_{per}$, loss function for protection $\mathcal{L}_{protect}$, the number of personalization loops $N_{per}$, the number of protection loops $N_{protect}$, learning rate in for personalization $\gamma_{per}$, learning rate in protection $\gamma_{protect}$. \\
\textbf{Output:} safeguarded model $\hat\theta$. \\
\textbf{Procedure:}
\begin{algorithmic}[1]
\State $j \gets 1, \theta_j \gets \theta$
\For {$j$ to $N_{protect}$} \Comment{Protection Path}
    \State $i \gets 1, \theta'_i \gets \theta_j$.copy(), $g \gets \varnothing$
    \For {$i$ to $N_{per}$} \Comment{Personalization Path}
    \State $\theta'_{i+1} \gets \theta'_{i} - \gamma_{per}\nabla_{\theta'_{i}}\mathcal{L}_{per}$ \Comment {Eq.~\eqref{eq:c_per2}}
    \State $g$.append($\nabla_{\theta'_{i+1}}\mathcal{L}_{protect}$)
    \EndFor
    \State $\nabla_{protect} \gets g$.sum() \Comment{Eq.~\eqref{eq:acc_grad}}
    \State $\theta_{j+1} \gets \theta_{j} - \gamma_{protect}\nabla_{protect}$ \Comment {Eq.~\eqref{eq:c_pro}}
\EndFor
\State \textbf{return} $\hat\theta \gets \theta_{N_{protect}}$
\end{algorithmic}
\end{algorithm}
\methodTwo{} involves two optimization paths: personalization and protection. 
The personalization path updates the model from the current protection state $\theta_j$ to intermediate state $\theta'_i$, using Eq.~\eqref{eq:l_per}:
\begin{align} 
\label{eq:c_per} 
    &\theta'_i = \theta_j, \\
    \label{eq:c_per2} 
    &\theta'_{i+1} = \theta'_i - \gamma_{per}\nabla_{\theta'_i}\mathcal{L}_{per},
\end{align}
where $\gamma_{per}$ is the learning rate for personalization, and $\theta'_{i+1}$ is the intermediate state at step $i+1$ during personalization. Using Eq.~\eqref{eq:c_per2}, we can simulate the future personalization trajectory via updating the model $\theta'_i$ iteratively, in the middle of protecting the DM.

For the protection path, we leverage these intermediate states acquired in the personalization path.
Specifically, we compute the gradient $\nabla_{i}$ of the model $\theta'_i$ with respect to $\mathcal{L}_{protect}$, at each state $i$ in the personalization path as follows:
\begin{equation}
    \nabla_{i} = \nabla_{\theta'_i}\mathcal{L}_{protect}.
\end{equation}
We then accumulate $\nabla_i$ during the whole personalization path (total of $N_{per}$ times) to compose a set of gradients, $g=\{\nabla_i\}^{N_{per}}_{i=1}$.
Using this set of gradients $g$, we can estimate the direction of protection from the summation of these accumulated gradients as follows:
\begin{align}
\label{eq:acc_grad}
    \nabla_{protect} = \sum^{N_{per}}_{i=1}\nabla_i.
\end{align}
Finally, we update the intermediate protection model $\theta_j$ with $\nabla_{protect}$ to obtain $\theta_{j+1}$: 
\begin{align}
\label{eq:c_pro} 
    &\theta_{j+1} = \theta_j - \gamma_{protect}\nabla_{protect}, 
\end{align}
where $\gamma_{protect}$ is the learning rate for protection.
By repeating this process for $N_{protect}$ times, we can obtain a safeguarded model $\hat\theta$, which is aware of the personalization path inherently for better protection. Algorithm~\ref{algorithm} illustrates the overall learning process of L2P for our APDM framework.

\vspace{-0.2cm}
\section{Experiments}
\vspace{-0.1cm}
\subsection{Experimental Setup}
\label{sec:5-1}
\vspace{-1mm}
\paragraph{Evaluation Metrics.}
To evaluate the effectiveness of APDM in protecting against personalization on specific subjects, we used two metrics: (i) the DINO score~\cite{caron2021emerging} as a similarity-based metric and (ii) BRISQUE~\cite{mittal2012no} for assessing image quality.
Additionally, we evaluated the preservation of the pre-trained model's generation capabilities by using (iii) the FID score \cite{heusel2017gans} for image quality, (iv) the CLIP score \cite{radford2021learning}, (v) TIFA~\cite{hu2023tifa}, and (vi) GenEval~\cite{ghosh2023geneval} for image-text alignment.

\begin{table}[t]
\caption{\textbf{Quantitative Comparison on Protection.} We measured the protection performance via DINO score \cite{caron2021emerging} and BRISQUE \cite{mittal2012no}. We examined the baseline on different number of clean images. If the number is 0, there are only perturbed images produced by data-poisoning approaches. The experiments were mainly conducted on two different subjects: person and dog.}
\label{tab:main}
\centering
\resizebox{0.9\linewidth}{!}{%
\begin{tabular}{lcccgccg}
\toprule
\multirow{2}{*}{Methods} &
  \multirow{2}{*}{\begin{tabular}[c]{@{}c@{}}\# Clean\\ Images\end{tabular}} &
  \multicolumn{3}{c}{DINO ($\downarrow$)} &
  \multicolumn{3}{c}{BRISQUE ($\uparrow$)} \\ \cmidrule{3-8}
                                 &     &\textit{``person''} &\textit{``dog''} & Avg. & \textit{``person''} & \textit{``dog''} & Avg. \\ \midrule
DreamBooth~\cite{ruiz2023dreambooth}& $N$&  0.6994  &    0.6056   & 0.6525 &  11.27  &    22.33   & 16.80 \\ \midrule
\multirow{3}{*}{AdvDM~\cite{liang2023adversarial}}&$0$& 0.5752  &   0.4247    & 0.4999 &  19.52   &    28.60   &  24.06    \\
                                 & $1$   & 0.5436   &   0.4393    & 0.4915 &  17.82   &   28.58    &   23.20   \\
                                 & $N-1$ & 0.6417   &   0.4775    & 0.5596 &  20.30   &   27.36    &  23.83    \\ \dashmidrule
\multirow{3}{*}{Anti-DreamBooth~\cite{van2023anti}} & $0$ &    0.5254   & \underline{0.4106} & 0.4680 &  \underline{26.90}  &   30.23    & \underline{28.56}     \\
                                 & $1$   & 0.6081   &    0.4704   & 0.5393 &  23.76  &   27.49    & 25.63     \\
                                 & $N-1$ & 0.6951   &    0.5304   & 0.6127 &  15.48  &    25.26   & 20.37     \\ \dashmidrule
\multirow{3}{*}{SimAC~\cite{Wang2024simac}}& $0$   &\underline{0.4448} & 0.4374& \underline{0.4411} & 23.73&    \underline{31.64}   & 27.69    \\
                                 & $1$   &  0.5824  &    0.4537   & 0.5181  &  18.04  &   29.54   & 23.79    \\
                                 & $N-1$ &  0.6991  &    0.5370   & 0.6181  &  14.28  &   27.05   & 20.67   \\ \dashmidrule
\multirow{3}{*}{PAP~\cite{wan2024prompt}}& $0$& 0.6556 & 0.5120   & 0.5838  &  22.61  &   30.20   & 26.41   \\
                                 & $1$   & 0.6690   & 0.5032      & 0.5861  &  22.02  &   29.00   & 25.51   \\
                                 & $N-1$ & 0.7028   & 0.5270      & 0.6149  &  19.64  &   23.41   & 21.53   \\ \midrule
\textbf{APDM (Ours)}             & $N$   &  \textbf{0.1375} &    \textbf{0.0959}   & \textbf{0.1167}  &\textbf{40.25} &   \textbf{60.74}    & \textbf{50.50}  \\ \bottomrule
\end{tabular}%
}
\vspace{-4mm}
\end{table}

\begin{figure}[t]
    \centering
    \includegraphics[width=\linewidth]{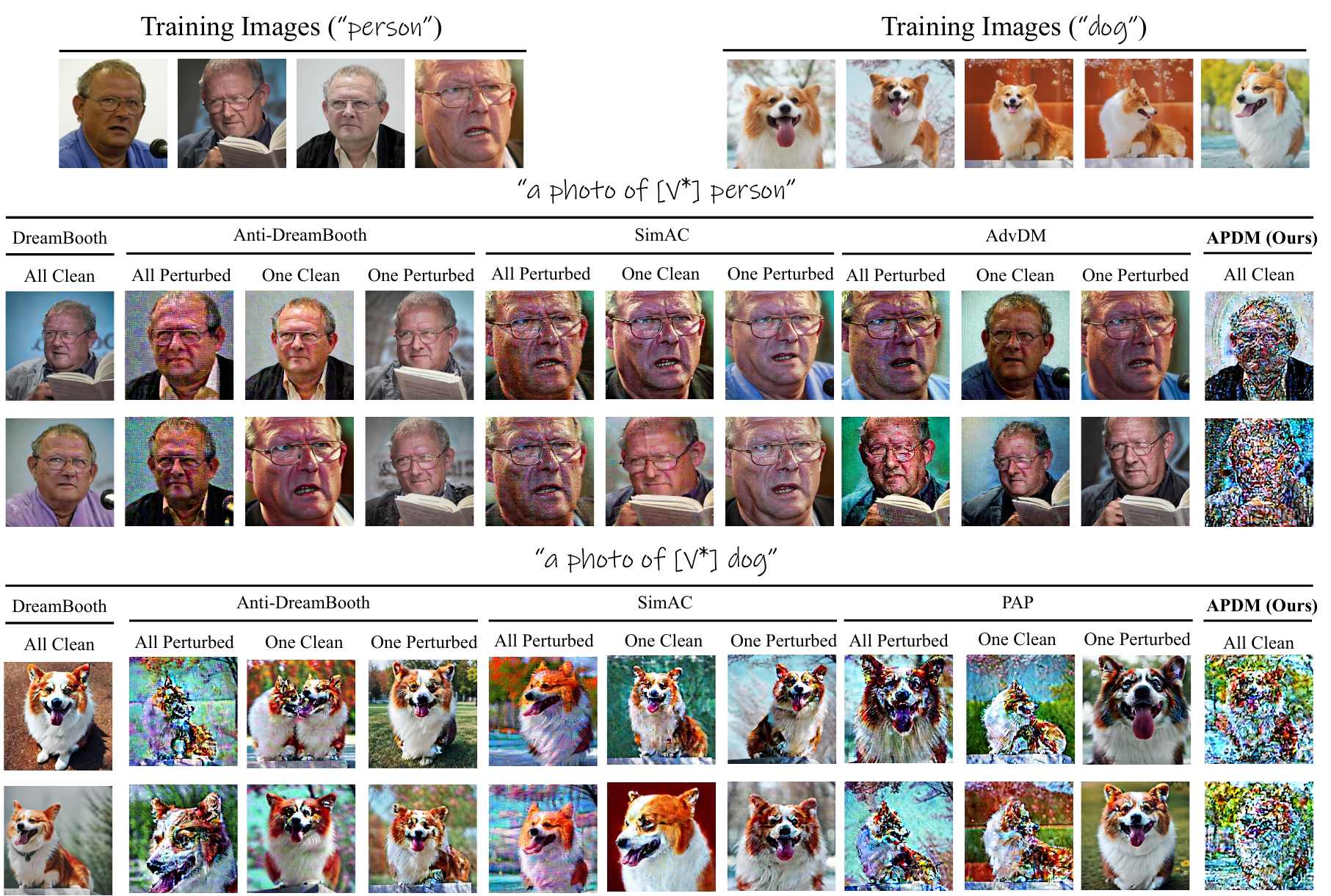}
    \caption{\textbf{Qualitative Comparison on Protection.} We examined the baselines and APDM on a protective aspect. We tested baselines on different circumstance - ``All Perturbed'', ``One Clean'', and ``One Perturbed''. In the ``All Perturbed'' setting, the baselines added perturbations to all training images. ``One Clean'' and ``One Perturbed'' settings are more difficult than ``All Perturbed'' setting, where the dataset contains one clean image or one perturbed image.}
    \label{fig:clean_all}
    \vspace{-0.5cm}
\end{figure}
\vspace{-0.3cm}
\paragraph{Baselines.}
We consider DreamBooth~\cite{ruiz2023dreambooth} and Custom Diffusion~\cite{kumari2023multi} as personalization methods.
The results of Custom Diffusion are presented in~\cref{supp:generalizablity}. 
For baselines, we include the previous protection approaches: (i) AdvDM \cite{xu2024perturbing}, (ii) Anti-DreamBooth~\cite{van2023anti}, (iii) SimAC~\cite{Wang2024simac}, and (iv) PAP~\cite{wan2024prompt}.
Following Anti-DreamBooth, we set the perturbation intensity for all baselines to 5e-2.

\vspace{-3mm}
\paragraph{Datasets.}
We used the datasets from both DreamBooth\footnote{https://github.com/google/dreambooth} \cite{ruiz2023dreambooth} and Anti-DreamBooth~\cite{van2023anti} to evaluate the protection performance.
The DreamBooth dataset contains 4-6 images per subject across various object classes such as dog, cat, and toy.
The Anti-DreamBooth dataset includes 4 images per person, consisting of facial images collected from CelebA-HQ~\cite{Karrs2018progressive} and VGGFace2~\cite{cao2018vggface2}.
To quantify the preservation performance of the model, we also used the MS-COCO 2014~\cite{lin2014microsoft} validation split.

\vspace{-3mm}
\paragraph{Implementation Details.}
We built APDM on Stable Diffusion 1.5 and Stable Diffusion 2.1~\cite{rombach2022high} with 512x512 resolution.
We used AdamW optimizer~\cite{loshchilov2019decoupled} with learning rates $\gamma_{per}=\gamma_{protect}=\mathrm{5e-6}$.
In DPO, we set the hyperparameter $\beta$ to 1.
In L2P, we used $N_{per}=20$ and $N_{protect}=800$.
We conducted all of our experiments on a single NVIDIA RTX A6000 GPU, and it took about 9 GPU hours to protect DM.
To synthesize images, we used PNDM scheduler~\cite{liu2022pseudo} with 20 steps.
For Stable Diffusion 2.1, we have attached the experimental results in~\cref{supp:generalizablity}.

\vspace{-1mm}
\subsection{Protection Performance}
\vspace{-1mm}
\label{sec:5-2}
As shown in Figure~\ref{fig:clean_all} and Table~\ref{tab:main}, we first evaluated the baselines and APDM from the perspective of protection.
We first personalized the pre-trained Stable Diffusion using DreamBooth~\cite{ruiz2023dreambooth} as a reference.
In this experiment, we considered three scenarios to test baselines and APDM.
For DreamBooth and APDM, only $N$ clean (\ie non-perturbed) images were used throughout the entire experiment (``All Clean'' in Figure~\ref{fig:clean_all}).
On the other hand, for data-poisoning baselines, we adopted different personalization scenarios.
For ``All Perturbed'' scenario, we utilized all perturbed images from each data-poisoning baseline.
Moreover, for ``One Clean'' scenario, we used $1$ clean image and $N-1$ perturbed images for personalization.
Lastly, the most challenging scenario, for ``One Perturbed'' scenario, there were only $1$ perturbed image and $N-1$ clean images in the dataset.

In Figure~\ref{fig:clean_all}, comparisons revealed their limitations as the scenarios become more challenging.
When only one perturbed image is used and the others remain clean, protection against personalization for the subjects becomes ineffective.
In contrast, despite the presence of clean images, APDM consistently demonstrated its robustness in more challenging scenarios (additional qualitative results in~\cref{supp:additional_qualitative}).
We also present a quantitative comparison in Table~\ref{tab:main}, highlighting that APDM outperforms data-poisoning approaches even under the most difficult conditions.
This is because APDM protects personalization at the model-level, making it robust to variations in the input data.
In addition, we also tested APDM in different scenarios (transform, such as flipping and blurring) and subjects such as ``cat'', ``sneaker'', ``glasses'', and ``clock'' (results in~\cref{supp:additional_experiments}).

\vspace{-2mm}
\subsection{Preservation Performance}
\vspace{-1mm}
\label{sec:5-3}
\begin{wraptable}{r}{0.65\linewidth}
  \begin{center}
  \begin{minipage}{\linewidth}
\vspace{-6mm}
\caption{\textbf{Preservation Performance on Image Quality and Image-Text Alignment.} We measured the image quality via FID score~\cite{heusel2017gans} and image-text alignment via CLIP score~\cite{radford2021learning}, TIFA~\cite{hu2023tifa}, and GenEval~\cite{ghosh2023geneval} on COCO 2014~\cite{lin2014microsoft} validation dataset. 
}
\label{tab:preservation_fid}
\centering
\resizebox{\linewidth}{!}{%
\begin{tabular}{lcccc}
\toprule
Methods          & FID ($\downarrow$) & CLIP ($\uparrow$) & TIFA ($\uparrow$) & GenEval ($\uparrow$) \\ \midrule
Stable Diffusion~\cite{rombach2022high} & 25.98 & 0.2878 & 78.76 & 0.4303 \\
\textbf{APDM (Ours)} & 28.85 & 0.2853 & 75.91 & 0.4017 \\ \bottomrule
\end{tabular}%
}
\vspace{-5mm}
    \end{minipage}
  \end{center}
\end{wraptable}
As described in Section~\ref{sec:4-3-2}, we updated the parameters of DM initialized with a pre-trained DM to obtain a safeguarded model.
To ensure its usability in future applications, it is essential to preserve the inherent capabilities of the pre-trained DMs during the protection process.
In this section, we evaluated the inherent performance based on image quality, image-text alignment of generated images, and the success of personalization for subjects not targeted by the protection.

\begin{table}[t]
\caption{\textbf{Preservation Performance on Personalization of Different Subjects.} We tried to personalize APDM to different subjects, such as ``cat'', ``sneaker'', and ``glasses''. We reported the personalization performance of DreamBooth \cite{ruiz2023dreambooth} these subjects as a reference.}
\label{tab:preservation_inter}
\centering
\resizebox{0.9\linewidth}{!}{%
\begin{tabular}{lcccgcccg}
\toprule
\multirow{2}{*}{Methods} &
  \multicolumn{4}{c}{DINO ($\uparrow$)} &
  \multicolumn{4}{c}{BRISQUE ($\downarrow$)} \\ \cmidrule(r){2-9} 
            & \textit{``cat''} & \textit{``sneaker''} & \textit{``glasses''} & Avg. & \textit{``cat''} & \textit{``sneaker''} & \textit{``glasses''} & Avg. \\ \midrule
DreamBooth~\cite{ruiz2023dreambooth}  & 0.4903 & 0.6110 & 0.6961 & 0.5991 & 25.32 & 23.14 & 19.01 & 22.49 \\ \dashmidrule
\textbf{APDM (Ours)} & 0.4231 & 0.7573 & 0.7198 & 0.6334 & 27.72 & 18.10 & 27.41 & 24.41 \\ \bottomrule
\end{tabular}%
}
\end{table}
Table~\ref{tab:preservation_fid} shows that APDM maintains high-quality image generation comparable to the pre-trained model. 
Beyond image quality and image-text alignment, we also evaluated its ability to personalize for different subjects using the protected DMs. 
Specifically, we tested personalization on models protected for ``person'' or ``dog'', using a new set of images featuring ``cat'', ``clock'' and ``glasses''.
As shown in Table~\ref{tab:preservation_inter}, these protected models remain effective for personalizing other subjects.
Overall, APDM successfully protects specific subjects while preserving personalization capabilities for others, making it suitable for handling diverse user requests in real-world applications.

\begin{table}[!t]

        \begin{minipage}{0.49\linewidth}
        \centering
\caption{\textbf{Ablation on the Effect of Image Pairing between $x^+_0$ and $x^-_0$.} We compared the protection performance with and without pairing.}
\label{tab:ablation_pair}
\centering
\resizebox{\linewidth}{!}{%
\begin{tabular}{ccccc}
\toprule
\multirow{2}{*}{Paired} & \multicolumn{2}{c}{DINO ($\downarrow$)} & \multicolumn{2}{c}{BRISQUE ($\uparrow$)} \\ \cmidrule{2-5}
 & \textit{``person''} & \textit{``dog''} & \textit{``person''} & \textit{``dog''} \\ \midrule
\xmark & 0.2770 & 0.3487 & 27.32 & 29.87 \\
\cmark & \textbf{0.1375} & \textbf{0.0959} & \textbf{40.25} & \textbf{60.74}  \\ \bottomrule
\end{tabular}%
}
        \end{minipage}\hfill
        \begin{minipage}{0.49\linewidth}
        \centering
\caption{\textbf{Ablation on the Effect of L2P.} We compared the performance between protection attempts without and with L2P.}
\label{tab:ablation_l2p}
\centering
\resizebox{\linewidth}{!}{%
\begin{tabular}{ccccc}
\toprule
\multirow{2}{*}{L2P} & \multicolumn{2}{c}{DINO ($\downarrow$)} & \multicolumn{2}{c}{BRISQUE ($\uparrow$)} \\ \cmidrule{2-5}
                 & \textit{``person''} & \textit{``dog''}  & \textit{``person''} & \textit{``dog''}  \\ \midrule
\xmark &  0.4454  & 0.3689 &  24.70   &  30.62 \\
\cmark &  \textbf{0.1375} &  \textbf{0.0959}  & \textbf{40.25}  & \textbf{60.74} \\ \bottomrule
\end{tabular}%
}
        \end{minipage} \\
        \begin{minipage}{0.49\linewidth}
        \centering
\caption{\textbf{Ablation on the Effect of $\beta$.} We compared the protection performance with different hyperparameter $\beta$.}
\label{tab:ablation_beta}
\centering
\resizebox{\linewidth}{!}{%
\begin{tabular}{ccccc}
\toprule
\multirow{2}{*}{$\beta$} & \multicolumn{2}{c}{DINO ($\downarrow$)} & \multicolumn{2}{c}{BRISQUE ($\uparrow$)} \\ \cmidrule{2-5} 
 & \textit{``person''} & \textit{``dog''} & \textit{``person''} & \textit{``dog''} \\ \midrule
 \textbf{1} & \textbf{0.1375} & \textbf{0.0959} & \textbf{40.25} & \textbf{60.74} \\
 10 & 0.5392 & 0.3885 & 13.58 & 15.14 \\
 100 & 0.5962 & 0.4755 & 12.21 & 14.10 \\ \bottomrule
\end{tabular}%
}
        \end{minipage}\hfill
        \begin{minipage}{0.49\linewidth}
        \centering
\caption{\textbf{Ablation on the Effect of $N_{per}$ in L2P.} We measured the performance in a protection aspect by varying $N_{per}$ of personalization path.}
\label{tab:ablation_iter}
\centering
\resizebox{\linewidth}{!}{%
\begin{tabular}{ccccc}
\toprule
\multirow{2}{*}{$N_{per}$} & \multicolumn{2}{c}{DINO ($\downarrow$)}     & \multicolumn{2}{c}{BRISQUE ($\uparrow$)}    \\ \cmidrule{2-5} 
 & \multicolumn{1}{c}{\textit{``person''}} & \multicolumn{1}{c}{\textit{``dog''}} & \multicolumn{1}{c}{\textit{``person''}} & \multicolumn{1}{c}{\textit{``dog''}} \\ \midrule
5                      & 0.3371 & 0.1923 & 37.89 & 39.48 \\
10                     & 0.2096 & 0.1342 & 38.14 & 47.15 \\
\textbf{20}            & \textbf{0.1375} & \textbf{0.0959} & \textbf{40.25} & \textbf{60.74} \\ \bottomrule
\end{tabular}%
}
        \end{minipage} \\

\end{table}
\subsection{Ablation Study}
\paragraph{Ablation on Loss Functions.}
In Section \ref{sec:4-3}, we introduced a novel objective, Direct Protective Optimization (DPO), which effectively prevents personalization while minimally degrading the model’s generation performance. 
In Table \ref{tab:ablation_pair}, we assessed the impact of pairing positive and negative images on protection performance. 
The results demonstrate that constructing image pairs significantly enhances performance by providing explicit guidance on which information should be encouraged or discouraged. 
Additionally, we investigated the effect of the hyperparameter $\beta$, which governs the strength of our DPO objective. 
As shown in Table~\ref{tab:ablation_beta}, our findings indicate that reducing $\beta$ allows APDM to more effectively prevent personalization.

\paragraph{Ablation on Optimization Scheme.}
\label{sec:5-4}
In Section \ref{sec:4-3-2}, we proposed a novel optimization scheme, Learning to Protect (L2P), which incorporates awareness of the personalization process during protection. 
In Table \ref{tab:ablation_l2p}, we compared the protection performance with and without L2P, and observed that incorporating the personalization trajectory significantly improves protection performance. 
Moreover, we examined the effect of the number of personalization paths ($N_{per}$). 
As shown in Table \ref{tab:ablation_iter}, increasing $N_{per}$ consistently improves performance. 
Despite this trend, we set $N_{per}=20$ as the default in our overall experiments, since it already achieved state-of-the-art performance.

\subsection{Additional Experiments}

As demonstrated in previous experiments, APDM effectively performs protection even in challenging cases, such as when clean images are used.
This robustness comes from its model-level defense mechanism, which allows protection to be achieved independently of the input data.
To further demonstrate this robustness, we examined whether APDM can also protect against perturbed data generated through data-poisoning methods.
Specifically, we generated perturbed data using Anti-DreamBooth~\cite{van2023anti} and evaluated APDM’s protection performance on these data.
As shown in Table~\ref{tab:perturbed_data}, APDM successfully prevents personalization even on perturbed data, confirming that its effectiveness is independent of the input variations.
\begin{table}[t]
\caption{\textbf{Protection performance of APDM on clean and perturbed data.} We evaluate whether APDM can maintain its protection capability regardless of input perturbations.}
\label{tab:perturbed_data}
\centering
\resizebox{0.7\linewidth}{!}{%
\begin{tabular}{lccc}
\toprule
Methods & {\begin{tabular}[c]{@{}c@{}}\# Clean\\ Images\end{tabular}} & DINO ($\downarrow$) & BRISQUE ($\uparrow$) \\ \midrule
DreamBooth~\cite{ruiz2023dreambooth}& $N$&  0.6869  &    16.69   \\ \dashmidrule
Anti-DreamBooth~\cite{van2023anti} & $0$ &  0.5646  & 22.50  \\ \midrule
\textbf{APDM (Ours)}             & $N$   &  \textbf{0.1375} &    \textbf{40.25} \\ 
\textbf{APDM (Ours, perturbed)}  & $0$   &  \underline{0.1702} &    \underline{40.20} \\ \bottomrule
\end{tabular}%
}
\end{table}

\begin{wraptable}{r}{0.6\linewidth}
  \begin{center}
  \begin{minipage}{\linewidth}
\vspace{-5mm}
\caption{\textbf{Protection performance of APDM under varying numbers of unseen images.} We evaluate whether APDM can maintain its protection capability across different input conditions and unseen data counts.}
\label{tab:unseen_data}
\centering
\resizebox{\linewidth}{!}{%
\begin{tabular}{lccc}
\toprule
Methods & \# of unseen & DINO ($\downarrow$) & BRISQUE ($\uparrow$) \\ \midrule
DreamBooth~\cite{ruiz2023dreambooth}    & $-$   &  0.6869 & 16.69    \\ \midrule
\multirow{4}{*}{\textbf{APDM (Ours)}}   & $-$   &  0.1375 & 40.25    \\ 
                                        & $4$   &  0.1616 & 38.14    \\
                                        & $8$   &  0.1994 &	38.87     \\ 
                                        & $12$  &  0.1873 &	38.87     \\ \bottomrule
\end{tabular}%
}
\vspace{-4mm}

    \end{minipage}
  \end{center}
\end{wraptable}
Building upon the previous analysis on perturbed data, we further investigated whether APDM maintains its protection capability when both the number and type of personalization data vary.
Specifically, this evaluation examined the generalization and scalability of APDM by considering two factors: (i) the use of unseen data that were not included during the protection stage, and (ii) the increased amount of personalization data per subject. 
As shown in Table~\ref{tab:unseen_data}, APDM consistently maintains protection performance even when 4–12 unseen images are introduced, confirming that its defense mechanism generalizes well to unseen samples and remains robust as the data volume increases.

To further assess the robustness of APDM under diverse personalization conditions, we additionally conducted experiments using varied text prompts and different unique identifiers, as well as an independent user study designed to evaluate real users’ preferences.
Due to the page limit, these extended results are provided in~\cref{supp:additional_experiments} (diverse prompt and identifier experiments) and~\cref{supp:user_study} (user study).
\vspace{-0.2cm}
\section{Conclusion}
\vspace{-1mm}
In this paper, we address privacy concerns in personalized DMs.
We highlight critical limitations of existing approaches, which depend on impractical assumptions (\eg exhaustive data poisoning) and fail to comply with privacy regulations.
Furthermore, we demonstrate that these approaches are easily circumvented when attackers use clean images or apply transformations to weaken the perturbation effects.
Therefore, we shifted the focus from data-centric defenses to model-level protection, aiming to directly prevent personalization through optimization rather than input modification.
To this end, we propose a novel framework APDM (Anti-Personalized Diffusion Models), which consists of a novel loss function, DPO (Direct Protective Optimization), and a new dual-path optimization scheme, L2P (Learning to Protect).
With APDM, we successfully prevented personalization while preserving the generative quality of the original model.
Experimental results demonstrate the effectiveness and robustness of APDM with promising outputs.
We hope our work extends the scope of anti-personalization towards more practical and appropriate real-world solutions.

\begin{ack}
This work was supported by Korea Planning \& Evaluation Institute of Industrial Technology (KEIT) grant funded by the Korea government (MOTIE) (RS-2024-00444344), and in part by Institute of Information \& communications Technology Planning \& Evaluation (IITP) grant funded by the Korea government (MSIT) under Grant No. RS2019-II190079 (Artificial Intelligence Graduate School Program (Korea University)), No. RS-2024-00457882 (AI Research Hub Project), and 2019-0-00004 (Development of Semi-Supervised Learning Language Intelligence Technology and Korean Tutoring Service for Foreigners).
Additionally, it was supported in part by the Institute of Information and Communications Technology Planning and Evaluation (IITP) Grant funded by the Korea Government (MSIT) ((Artificial Intelligence Innovation Hub) under Grant RS-2021-II212068).
\end{ack}

\medskip

{
\small
\bibliographystyle{splncs04}
\bibliography{main}
}


\newpage
\appendix

In this appendix, we provide detailed proofs and derivations, and additional experimental results that were not included in the main paper due to page limits.
The contents of the appendix are as follows:
\begin{itemize}
    \item \cref{supp:proof_derivation}: Derivation and proof of Proposition 1, Theorem 1, and Direct Protective Optimization (DPO) objective.
    \item \cref{supp:additional_experiments}: Additional experiments including empirical results about na\"ive approach, comparison on protection with image transformations, and protection performance for different subjects. 
    \item \cref{supp:generalizablity}: Generalizability of APDM on different personalization methods, Stable Diffusion version, unique identifier, and diverse test prompts.
    \item \cref{supp:user_study}: User study about protection performance.
    \item \cref{supp:additional_qualitative}: Additional qualitative results extending to the experimental results in the main paper.
    \item \cref{supp:add_motivate}: Additional explanation about our motivation. 
    \item \cref{supp:limitation}: The limitations and broader impacts of APDM, and a discussion of future work.
\end{itemize}

\section{Proofs and Derivation}
\label{supp:proof_derivation}

In this section, we present the formal proofs and derivations supporting our main theoretical contributions discussed in the main paper.
We begin by providing a rigorous proof for Proposition~\ref{prop:convergence condition} (Appendix~\ref{supp:proof_prop_1}), followed by the complete proof for Theorem~\ref{theo:theorem1} (Appendix~\ref{supp:proof_theo_1}).
Subsequently, we detail the step-by-step derivation of our proposed DPO loss function in Appendix~\ref{supp:derivation}.

\subsection{Proof of Proposition 1}
\label{supp:proof_prop_1}
The primary goal of Proposition~\ref{prop:convergence condition} is to identify and establish the necessary conditions under which na\"ive approach converges.
We begin the proof by recalling the loss function of na\"ive approach, Equation~\eqref{eq:adv} in our main paper:
\begin{align} 
\label{eq:L_adv} 
    \mathcal{L}_{adv}=-\mathcal{L}^{per}_{simple} + \lambda\mathcal{L}_{ppl},
\end{align}
where $\lambda$ is positive scalar ($\lambda>0$) to weight the $\mathcal{L}_{ppl}$, and each term is defined as follows:
\begin{align}
\label{eq:L_simple_sup}
    \mathcal{L}^{per}_{simple}&=\mathbb{E}_{x_0,t,c,\epsilon\sim\mathcal{N}(0,I)}{\Vert \epsilon_\theta(x_t,t,c) - \epsilon\Vert}^2_2, \\
\label{eq:L_ppl_sup}
    \mathcal{L}_{ppl}&=\mathbb{E}_{x^{pr}_0,t,c^{pr},\epsilon\sim\mathcal{N}(0,I)}{\Vert \epsilon_\theta(x^{pr}_t,t,c^{pr}) - \epsilon\Vert}^2_2.
\end{align}
In optimization theory, a fundamental necessary condition for a differentiable function to attain a local minimum is that its \textit{derivative with respect to the optimization variables must be zero}.
This is often referred to as the first-order necessary condition for optimality. 
Applying this principle to our case, for $\mathcal{L}_{adv}$ to converge to a stable point with respect to the model parameters $\theta$, the derivative must be zero as:
\begin{align}
\label{eq:conv_cond}
    \nabla_\theta \mathcal{L}_{adv} = 0.
\end{align}
To address the condition in Equation~\eqref{eq:conv_cond}, we compute the gradient of $\mathcal{L}_{adv}$ with respect to $\theta$.
To simplify the computation, we first recall the MSE loss term as:
\begin{align}
\label{eq:simple_mse}
    \Vert u-v \Vert^2_2 = u^\top u-2u^\top v+v^\top v.
\end{align}
Using the expansion of Equation~\eqref{eq:simple_mse}, we can rewrite the MSE terms of Equation~\eqref{eq:L_simple_sup} and Equation~\eqref{eq:L_ppl_sup} as follows:
\begin{align}
\label{eq:expand_per_in}
    \Vert \epsilon_\theta(x_t,t,c) - \epsilon\Vert^2_2 &={\epsilon_\theta^{per}}^\top{\epsilon_\theta^{per}}-2{\epsilon_\theta^{per}}^\top\epsilon+\epsilon^\top\epsilon, \\
\label{eq:expand_ppl_in}
    \Vert \epsilon_\theta(x^{pr}_t,t,c^{pr}) - \epsilon\Vert^2_2 &={\epsilon_\theta^{ppl}}^\top{\epsilon_\theta^{ppl}}-2{\epsilon_\theta^{ppl}}^\top\epsilon+\epsilon^\top\epsilon,
\end{align}
where $\epsilon_\theta^{per}=\epsilon_\theta(x_t,t,c)$ and $\epsilon_\theta^{ppl}=\epsilon_\theta(x^{pr}_t,t,c^{pr})$.
For notational simplicity in the subsequent derivations, we will omit the input variables (\eg $x_t, t, c$) and use the superscripts \textit{per} and \textit{ppl} to distinguish both terms in $\mathcal{L}_{simple}^{per}$ and $\mathcal{L}_{ppl}$ respectively.
Now, substituting the expanded forms from Equation~\eqref{eq:expand_per_in} and Equation~\eqref{eq:expand_ppl_in} back into $\mathcal{L}_{simple}^{per}$ and $\mathcal{L}_{ppl}$, we can rewrite these.

For $\mathcal{L}_{simple}^{per}$, using Equation~\eqref{eq:expand_per_in}:
\begin{align}
    \mathcal{L}_{simple}^{per}=\mathbb{E}_{x_0,t,c,\epsilon\sim\mathcal{N}(0,I)}{[{\epsilon_\theta^{per}}^\top{\epsilon_\theta^{per}}-2{\epsilon_\theta^{per}}^\top\epsilon+\epsilon^\top\epsilon]}.
\end{align}
Moreover, by the linearity of expectation, which includes the property $\mathbb{E}[A+B]=\mathbb{E}[A]+\mathbb{E}[B]$ (additivity principle), we can distribute expectation as follows:
\begin{align}
\label{eq:expand_per}
    \mathcal{L}_{simple}^{per}=\mathbb{E}[{\epsilon_\theta^{per}}^\top{\epsilon_\theta^{per}}]-2\mathbb{E}[{\epsilon_\theta^{per}}^\top\epsilon]+\mathbb{E}[\epsilon^\top\epsilon].
\end{align}
Please note that for readability, we also omit the explicit subscript variables of the expectation.

Similarly, for $\mathcal{L}_{ppl}$, using Equation~\eqref{eq:expand_ppl_in} and linearity of expectation:
\begin{align}
\begin{split}
\label{eq:expand_ppl}
    \mathcal{L}_{ppl}&=\mathbb{E}_{x^{pr}_0,t,c^{pr},\epsilon\sim\mathcal{N}(0,I)}{[{\epsilon_\theta^{ppl}}^\top{\epsilon_\theta^{ppl}}-2{\epsilon_\theta^{ppl}}^\top\epsilon+\epsilon^\top\epsilon]} \\
    &=\mathbb{E}[{\epsilon_\theta^{ppl}}^\top{\epsilon_\theta^{ppl}}]-2\mathbb{E}[{\epsilon_\theta^{ppl}}^\top\epsilon]+\mathbb{E}[\epsilon^\top\epsilon].
\end{split}
\end{align}

These expanded expressions (Equation~\eqref{eq:expand_per} and Equation~\eqref{eq:expand_ppl}) simplify the subsequent gradient derivations.
To compute the gradients of these loss functions, we will differentiate the terms within the expectation, which is permissible under suitable regularity conditions by applying the \textit{Leibniz Rule}.
We first consider the derivatives of the core components that appear inside the expectations, with respect to the model parameters $\theta$.
\begin{align}
    &\nabla_\theta(\epsilon_\theta^\top\epsilon_\theta)=2J_\theta^\top \epsilon_\theta,  \\
    &\nabla_\theta(\epsilon_\theta^\top\epsilon) = J_\theta^\top \epsilon, \\
    &\nabla_\theta(\epsilon^\top\epsilon) = 0,
\end{align}
where $J_\theta=\frac{\partial}{\partial\theta}\epsilon_\theta$ and $\epsilon$ is independent of $\theta$, the derivative of any term solely dependent on $\epsilon$ (\ie $\epsilon^\top\epsilon$) with respect to $\theta$ is zero.
Using these results, we can now express the gradient of MSE loss term inside the expectations. 
For the term in $\mathcal{L}_{simple}^{per}$:
\begin{align}
\begin{split}
\label{eq:mse_per_dev}
    \nabla_\theta\Vert \epsilon_\theta(x_t,t,c) - \epsilon\Vert^2_2&=\frac{\partial}{\partial\theta}({\epsilon_\theta^{per}}^\top{\epsilon_\theta^{per}})-2\frac{\partial}{\partial\theta}({\epsilon_\theta^{per}}^\top\epsilon)+\frac{\partial}{\partial\theta}(\epsilon^\top\epsilon) \\
    &=2{J_\theta^{per}}^\top\epsilon_\theta-2{J_\theta^{per}}^\top\epsilon.
\end{split} 
\end{align}
And for the term in $\mathcal{L}_{ppl}$:
\begin{align}
\begin{split}
\label{eq:mse_ppl_dev}
    \nabla_\theta\Vert \epsilon_\theta(x^{pr}_t,t,c^{pr}) - \epsilon\Vert^2_2&=\frac{\partial}{\partial\theta}({\epsilon_\theta^{ppl}}^\top{\epsilon_\theta^{ppl}})-2\frac{\partial}{\partial\theta}({\epsilon_\theta^{ppl}}^\top\epsilon)+\frac{\partial}{\partial\theta}(\epsilon^\top\epsilon) \\
    &=2{J_\theta^{ppl}}^\top\epsilon_\theta-2{J_\theta^{ppl}}^\top\epsilon.
\end{split}
\end{align}
Since $\mathcal{L}_{simple}^{per}$ and $\mathcal{L}_{ppl}$ are expectations of the terms whose gradients were derived in Equation~\eqref{eq:mse_per_dev} and Equation~\eqref{eq:mse_ppl_dev}, and we apply the \textit{Leibniz Rule}.
This allows us to take the expectation of those gradients to find the final gradients of the loss functions:
\begin{align}
\label{eq:grad_per}
    \nabla_\theta\mathcal{L}_{simple}^{per}&=2\mathbb{E}[{J_\theta^{per}}^\top\epsilon]-2\mathbb{E}[{J_\theta^{per}}^\top\epsilon], \\
\label{eq:grad_ppl}
    \nabla_\theta\mathcal{L}_{ppl}&=2\mathbb{E}[{J_\theta^{ppl}}^\top\epsilon]-2\mathbb{E}[{J_\theta^{ppl}}^\top\epsilon].
\end{align}

Consequently, using the gradients (Equation~\eqref{eq:grad_per} and Equation~\eqref{eq:grad_ppl}), we can determine the convergence condition for $\mathcal{L}_{adv}$ with respect to $\theta$ as:
\begin{align}
\begin{split}
\label{eq:grad_adv}
    \nabla_\theta\mathcal{L}_{adv}&=-\nabla_\theta\mathcal{L}_{simple}^{per}+\lambda\nabla_\theta\mathcal{L}_{ppl} \\
    &=-\{2\mathbb{E}[{J_\theta^{per}}^\top\epsilon]-2\mathbb{E}[{J_\theta^{per}}^\top\epsilon]\}+\lambda\{2\mathbb{E}[{J_\theta^{ppl}}^\top\epsilon]-2\mathbb{E}[{J_\theta^{ppl}}^\top\epsilon]\}.
\end{split}
\end{align}
Based on Equation~\eqref{eq:conv_cond}, rearranging Equation~\eqref{eq:grad_adv} yields the final condition for Proposition~\ref{prop:convergence condition}:
\begin{align}
\label{eq:prop1}
    \nabla_\theta\mathcal{L}_{simple}^{per}=\lambda\nabla_\theta\mathcal{L}_{ppl}.
\end{align}
The result in Equation~\eqref{eq:prop1} indicates that for $\mathcal{L}_{adv}$ to converge, the gradients of 
$\mathcal{L}_{simple}^{per}$ and $\mathcal{L}_{ppl}$ must point in the same direction, as $\lambda>0$.
This completes the proof of Proposition~\ref{prop:convergence condition}.

\subsection{Proof of Theorem 1}
\label{supp:proof_theo_1}

In Proposition~\ref{prop:convergence condition}, we establish that for $\mathcal{L}_{adv}$ to converge, a necessary condition is $\nabla_\theta\mathcal{L}_{simple}^{per}=\lambda\nabla_\theta\mathcal{L}_{ppl}$.
Based on this proposition, we now prove Theorem~\ref{theo:theorem1}.
Our proof will demonstrate how the aforementioned convergence condition (Equation~\eqref{eq:prop1}) inherently conflicts with the goal of simultaneously decreasing both $-\mathcal{L}_{simple}^{per}$ and $\mathcal{L}_{ppl}$.

To figure this out, we analyze the impact of a parameter update, $\Delta\theta$, on each loss term using a first-order \textit{Taylor Expansion}.
A parameter update $\Delta\theta$ derived from a gradient descent step on $\mathcal{L}_{adv}$, and assuming the scalar $\lambda=1$ for simplicity in this derivation.
$\Delta\theta$ can be defined as:
\begin{align}
    \Delta\theta = -\eta\frac{\partial}{\partial\theta}\mathcal{L}_{adv},
\end{align}
where $\eta>0$ is the learning rate.
The change in $\mathcal{L}_{simple}^{per}$ due to $\Delta\theta$ can be approximated by first-order \textit{Taylor Expansion}:
\begin{align}
\begin{split}
\label{eq:per_diff}
    \mathcal{L}_{simple}^{per}(\theta+\Delta\theta)-\mathcal{L}_{simple}^{per}(\theta)&\approx [\frac{\partial}{\partial\theta}\mathcal{L}_{simple}^{per}(\theta)]^\top\Delta\theta \\
    &\approx [\frac{\partial}{\partial\theta}\mathcal{L}_{simple}^{per}(\theta)]^\top\{-\eta[-\frac{\partial}{\partial\theta}\mathcal{L}_{simple}^{per}(\theta)+\frac{\partial}{\partial\theta}\mathcal{L}_{ppl}(\theta)]\} \\
    &\approx \eta\Vert\frac{\partial}{\partial\theta}\mathcal{L}_{simple}^{per}(\theta) \Vert^2-\eta[\frac{\partial}{\partial\theta}\mathcal{L}_{simple}^{per}(\theta)]^\top[\frac{\partial}{\partial\theta}\mathcal{L}_{ppl}(\theta)],
\end{split}
\end{align}
where $\mathcal{L}(\theta)$ means the the loss calculated with the parameter $\theta$.
Our objective is to minimize $-\mathcal{L}_{simple}^{per}$, which is equivalent to increasing $\mathcal{L}_{simple}^{per}$.
For this reason, the difference of $\mathcal{L}_{simple}^{per}$ in Equation~\eqref{eq:per_diff} is greater than zero. 
Using this condition, we can obtain the final inequality from Equation~\eqref{eq:per_diff} as:
\begin{align}
\label{eq:ineq_per}
    \Vert\frac{\partial}{\partial\theta}\mathcal{L}_{simple}^{per}(\theta) \Vert^2 > [\frac{\partial}{\partial\theta}\mathcal{L}_{simple}^{per}(\theta)]^\top[\frac{\partial}{\partial\theta}\mathcal{L}_{ppl}(\theta)].
\end{align}
This inequality (Equation~\eqref{eq:ineq_per}) represents the condition under which the parameter update leads to an increase in $\mathcal{L}_{simple}^{per}$.
Similarly, we derive the impact of the parameter update $\Delta\theta$ on $\mathcal{L}_{ppl}$.
\begin{align}
\begin{split}
\label{eq:ppl_diff}
    \mathcal{L}_{ppl}(\theta+\Delta\theta)-\mathcal{L}_{ppl}(\theta)&\approx [\frac{\partial}{\partial\theta}\mathcal{L}_{ppl}(\theta)]^\top\Delta\theta \\
    &\approx [\frac{\partial}{\partial\theta}\mathcal{L}_{ppl}(\theta)]^\top\{-\eta[-\frac{\partial}{\partial\theta}\mathcal{L}_{simple}^{per}(\theta)+\frac{\partial}{\partial\theta}\mathcal{L}_{ppl}(\theta)]\} \\
    &\approx \eta[\frac{\partial}{\partial\theta}\mathcal{L}_{ppl}(\theta)]^\top[\frac{\partial}{\partial\theta}\mathcal{L}_{simple}^{per}(\theta)]-\eta\Vert\frac{\partial}{\partial\theta}\mathcal{L}_{ppl}(\theta) \Vert^2.
\end{split}
\end{align}
To minimize $\mathcal{L}_{ppl}$, it should decrease, which means that the change approximated by Equation~\eqref{eq:ppl_diff} must be less than zero.
We can also derive the condition as:
\begin{align}
\label{eq:ineq_ppl}
    [\frac{\partial}{\partial\theta}\mathcal{L}_{ppl}(\theta)]^\top[\frac{\partial}{\partial\theta}\mathcal{L}_{simple}^{per}(\theta)]<\Vert\frac{\partial}{\partial\theta}\mathcal{L}_{ppl}(\theta) \Vert^2.
\end{align}

Equation~\eqref{eq:ineq_per} and Equation~\eqref{eq:ineq_ppl} share the common inner product term $[\frac{\partial}{\partial\theta}\mathcal{L}_{ppl}(\theta)]^\top[\frac{\partial}{\partial\theta}\mathcal{L}_{simple}^{per}(\theta)]$.
Recall from Proposition~\ref{prop:convergence condition}, for converging $\mathcal{L}_{adv}$, the gradients of the two terms must point in the same direction.
This relationship allows us to remove the cosine term in the inner product ($\because \cos(0)=1$). 
Based on this, we can rewrite the inequalities as below:
\begin{align}
\label{eq:ineq_per_cos}
     |\frac{\partial}{\partial\theta}\mathcal{L}_{simple}^{per}(\theta)|\cdot|\frac{\partial}{\partial\theta}\mathcal{L}_{ppl}(\theta)|&<\Vert\frac{\partial}{\partial\theta}\mathcal{L}_{simple}^{per}(\theta) \Vert^2, \\
\label{eq:ineq_ppl_cos}
    |\frac{\partial}{\partial\theta}\mathcal{L}_{ppl}(\theta)|\cdot|\frac{\partial}{\partial\theta}\mathcal{L}_{simple}^{per}(\theta)|&<\Vert\frac{\partial}{\partial\theta}\mathcal{L}_{ppl}(\theta) \Vert^2.
\end{align}
We can further rearrange the above inequality as:
\begin{align}
\label{eq:the1_per}
    |\frac{\partial}{\partial\theta}\mathcal{L}_{ppl}(\theta)|<|\frac{\partial}{\partial\theta}\mathcal{L}_{simple}^{per}(\theta)|, \\
\label{eq:the1_ppl}
    |\frac{\partial}{\partial\theta}\mathcal{L}_{ppl}(\theta)|>|\frac{\partial}{\partial\theta}\mathcal{L}_{simple}^{per}(\theta)|.
\end{align}
This rearrangement relies on the assumption that both individual gradients are non-zero, \ie $|\frac{\partial}{\partial\theta}\mathcal{L}_{simple}^{per}(\theta)|>0$ and $|\frac{\partial}{\partial\theta}\mathcal{L}_{ppl}(\theta)|>0$.
This assumption holds for any $\theta$ that is not already a local optimum for both individual objectives.

Recall the objectives: to increase $\mathcal{L}_{simple}^{per}$, the condition in Equation~\eqref{eq:the1_per} must hold for the current parameter update.
On the other hand, to decrease $\mathcal{L}_{ppl}$, the condition in Equation~\eqref{eq:the1_ppl} must satisfy for the same parameter update.
These two conditions are mutually exclusive.
This contradiction demonstrates that if the system is at a point satisfying the convergence condition for $\mathcal{L}_{adv}$ (Proposition 1), the objective of simultaneously decreasing $-\mathcal{L}_{simple}^{per}$ and $\mathcal{L}_{ppl}$ cannot be achieved. 
Therefore, the na\"ive approach $\mathcal{L}_{adv}$, as composed of these conflicting objectives under its own convergence condition, generally fails to converge to a point that effectively optimizes both. 
This completes the proof of Theorem~\ref{theo:theorem1}.

\subsection{Derivation of Objective}
\label{supp:derivation}
Starting from Equation~\eqref{eq:eq13} in the main paper, the loss function for the reward function $r(\cdot)$ can be expressed as:
\begin{align}
    \mathcal{L}_r=-\mathbb{E}_{x^+_0,x^-_0} \log\sigma(r(x^+_0)-r(x^-_0)).
    \label{supp:eq_reward_objective}
\end{align}
Reinforcement Learning from Human Feedback (RLHF) aims to maximize the distribution $p_\theta(x_0)$ under regularization using KL-divergence:
\begin{align}
    \max_{p_\theta}\mathbb{E}_{x_0}r(x_0)-\beta D_{KL}(p_\theta(x_0) \Vert p_\phi(x_0)),
    \label{supp:eq_rlhf}
\end{align}
where $\phi$ is reference distribution.
From Equation~\eqref{supp:eq_rlhf}, we can obtain a unique solution $p_\theta^*(x_0)$:
\begin{align}
    p_\theta^*(x_0)=p_\phi(x_0)\exp(r(x_0)/\beta)/Z,
    \label{supp:eq_partition}
\end{align}
where $Z=\sum_{x_0}p_\phi(x_0)\exp(r(x_0)/\beta)$ is the partition function.
The reward function can be re-written using Equation~\eqref{supp:eq_partition}:
\begin{align}
    r(x_0)=\beta\log\cfrac{p_\theta^*(x_0)}{p_\phi(x_0)}+\beta\log Z.
    \label{supp:eq_rewritten_reward}
\end{align}
From Equation~\eqref{supp:eq_reward_objective} and Equation~\eqref{supp:eq_rewritten_reward}, the reward objective is:
\begin{align}
    \mathcal{L}_{r}=-\mathbb{E}_{x^+_0,x^-_0}[\log\sigma(\beta\log\cfrac{p_\theta^*(x^+_0)}{p_\phi(x^+_0)}-\beta\log\cfrac{p_\theta^*(x^-_0)}{p_\phi(x^-_0)})].
    \label{supp:eq_loss_dpo_pre}
\end{align}
However, this objective cannot directly applied to diffusion models since the parameterized distribution $p_\theta(x_0)$ is intractable.
Therefore, Diffusion-DPO~\cite{wallace2024diffusion} introduces the latents $x_{1:T}$ to consider possible diffusion paths from $x_T$ to $x_0$, and re-defines the reward function as follows:
\begin{align}
    r(x_0)=\mathbb{E}_{p_\theta(x_{1:T}\vert x_0)}R(x_0).
    \label{supp:eq_new_reward_objective}
\end{align}
Following Equation~\eqref{supp:eq_new_reward_objective}, Equation~\eqref{supp:eq_rlhf} can also be written as follows:
\begin{align}
    \max_{p_\theta}\mathbb{E}_{x_{0:T}\sim p(x_{0:T})}r(x_0)-\beta D_{KL}(p_\theta(x_{0:T}) \Vert p_\phi(x_{0:T})).
\end{align}
Similar to the expansion from Equation~\eqref{supp:eq_rlhf} to Equation~\eqref{supp:eq_loss_dpo_pre}, we can obtain the reward objective as:
\begin{align}
    \begin{split}
        \mathcal{L}_{r}=-\mathbb{E}_{x^+_0,x^-_0}[\log\sigma\{\mathbb{E}_{p_\theta(x^+_{1:T}\vert x^+_0),p_\theta(x^-_{1:T}\vert x^-_0)}
        (\beta\log\cfrac{p_\theta^*(x^+_{0:T})}{p_\phi(x^+_{0:T})}-\beta\log\cfrac{p_\theta^*(x^-_{0:T})}{p_\phi(x^-_{0:T})})\}].
    \end{split}
    \label{supp:eq_loss_dpo}
\end{align}
Since $-\log\sigma(\cdot)$ is a convex function, we can leverage \textit{Jensen's inequality}:
\begin{align}
    \begin{split}
        \mathcal{L}_r&\le-\mathbb{E}_{x^+_0,x^-_0,p_\theta(x^+_{1:T}\vert x^+_0),p_\theta(x^-_{1:T}\vert x^-_0)} \\
        &[\log\sigma\{\beta\log\cfrac{p_\theta^*(x^+_{0:T})}{p_\phi(x^+_{0:T})}-\beta\log\cfrac{p_\theta^*(x^-_{0:T})}{p_\phi(x^-_{0:T})}\}].
    \end{split}
\end{align}
Note that $p_\theta(x_{1:T}\vert x_0)$ is intractable.
Therefore, we utilize $q(x_{1:T}|x_0)$ to approximate $p_\theta(x_{1:T}\vert x_0)$:
\begin{align}
    \begin{split}
        \mathcal{L}_r&\le-\mathbb{E}_{x^+_0,x^-_0,q(x^+_{1:T}\vert x^+_0),q(x^-_{1:T}\vert x^-_0)}[\log\sigma\{\beta\log\cfrac{p_\theta^*(x^+_{0:T})}{p_\phi(x^+_{0:T})}-\beta\log\cfrac{p_\theta^*(x^-_{0:T})}{p_\phi(x^-_{0:T})}\}].
    \end{split}
\end{align}
Since $p_\theta(x_{0:T})=p_\theta(x_T)\prod^\top_{t=1}p_\theta(x_{t-1}\vert x_t)\ $ can be expressed as a Markov chain, we can derive the above equation as:
\begin{align}
    \begin{split}
        \mathcal{L}_r&\le-\mathbb{E}_{x^+_0,x^-_0,q(x^+_{1:T}\vert x^+_0),q(x^-_{1:T}\vert x^-_0)}\\
        &[\log\sigma\{\beta\sum^\top_{t=1}\log\cfrac{p_\theta^*(x^+_{t-1}\vert x^+_t)}{p_\phi(x^+_{t-1}\vert x^+_t)}-\log\cfrac{p_\theta^*(x^-_{t-1}\vert x^-_t)}{p_\phi(x^-_{t-1}\vert x^-_t)}\}], \\
        &=-\mathbb{E}_{x^+_0,x^-_0,q(x^+_{1:T}\vert x^+_0),q(x^-_{1:T}\vert x^-_0)}\\
        &[\log\sigma\{\beta\sum^\top_{t=1}(\log\cfrac{p^*_\theta(x^+_{t-1}\vert x^+_t)}{q(x^+_{t-1}\vert x^+_t)}-\log\cfrac{p_\phi(x^+_{t-1}\vert x^+_t)}{q(x^+_{t-1}\vert x^+_t)})\\
        &\quad\quad\quad\quad\quad\quad-(\log\cfrac{p^*_\theta(x^-_{t-1}\vert x^-_t)}{q(x^-_{t-1}\vert x^-_t)}-\log\cfrac{p_\phi(x^-_{t-1}\vert x^-_t)}{q(x^-_{t-1}\vert x^-_t)})\}], \\
        &=-\mathbb{E}_{x^+_0,x^-_0,q(x^+_{1:T}\vert x^+_0),q(x^-_{1:T}\vert x^-_0)}\\
        &[\log\sigma\{\beta\sum^\top_{t=1}(D_{KL}(q(x^+_{t-1}\vert x^+_t)\Vert p^*_\theta(x^+_{t-1}\vert x^+_t))\\
        &\quad\quad\quad\quad\quad\quad-D_{KL}(q(x^+_{t-1}\vert x^+_t) \Vert p_\phi(x^+_{t-1}\vert x^+_t)))\\
        &\quad\quad\quad\quad\quad\quad-(D_{KL}(q(x^-_{t-1}\vert x^-_t)\Vert p^*_\theta(x^-_{t-1}\vert x^-_t))\\
        &\quad\quad\quad\quad\quad\quad-D_{KL}(q(x^-_{t-1}\vert x^-_t) \Vert p_\phi(x^-_{t-1}\vert x^-_t)))\}].
    \end{split}
\end{align}
By leveraging ELBO, we can obtain our final objective, Equation~\eqref{eq:eq14} in the main paper:
\begin{align}
    \begin{split}
        \mathcal{L}_{DPO}=&-\mathbb{E}_{x^+_0,x^-_0,c,t,\epsilon \sim N(0,I)}\\
        &\log \sigma(-\beta((\Vert\epsilon_\theta(x^+_t,t,c)-\epsilon\Vert^2_2\\
        &\quad\quad\quad\quad-\Vert\epsilon_\phi(x^+_t,t,c)-\epsilon\Vert^2_2)\\
        &\quad\quad\quad\quad-(\Vert\epsilon_\theta(x^-_t,t,c)-\epsilon\Vert^2_2\\
        &\quad\quad\quad\quad-\Vert\epsilon_\phi(x^-_t,t,c)-\epsilon\Vert^2_2))).
    \end{split}
\end{align}

\begin{figure}[!h]
    \centering
    \vspace{-4mm}
    \includegraphics[width=\linewidth]{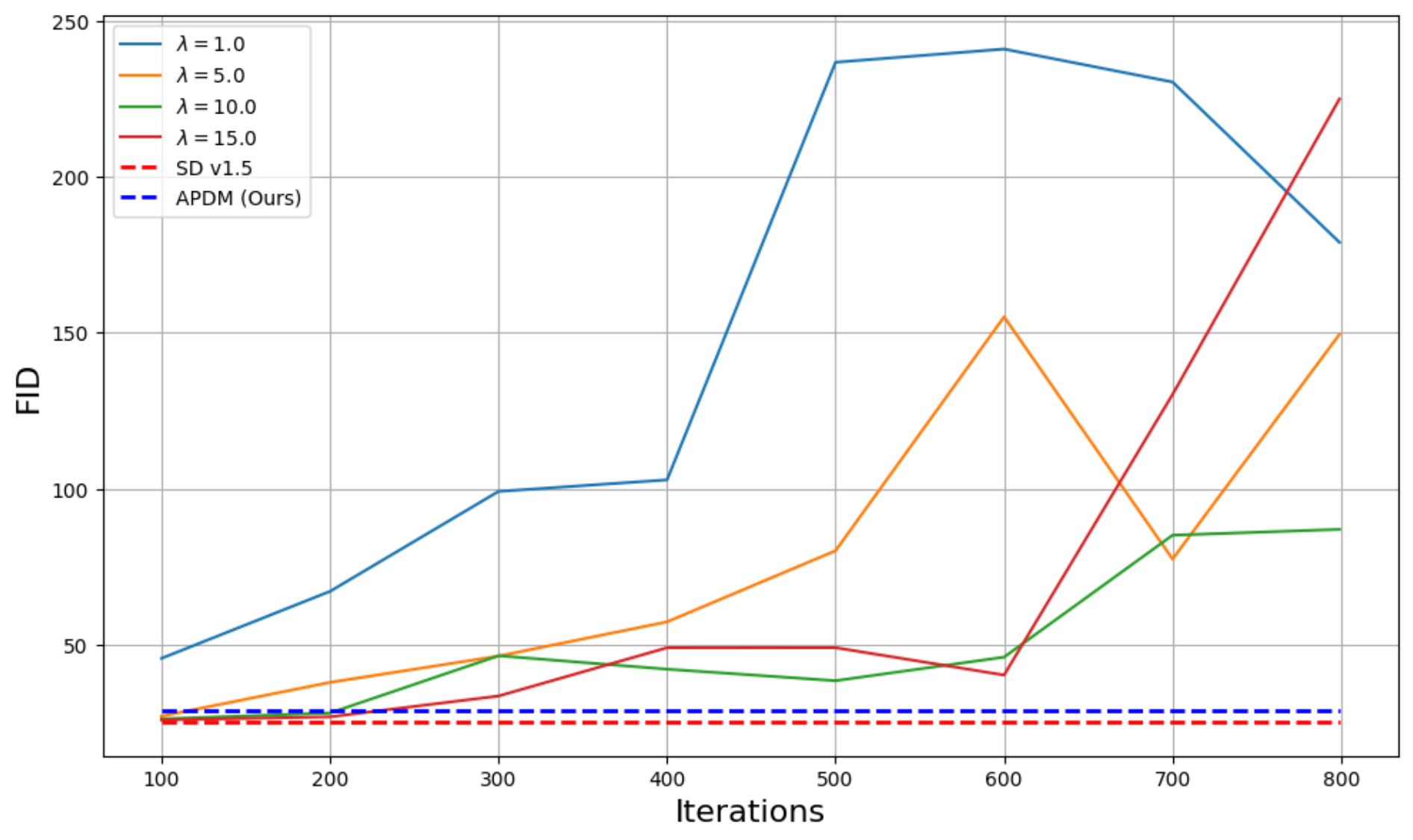}
    \caption{\textbf{FID variation during the training with $\mathcal{L}_{adv}.$} We measured the image quality via FID score~\cite{heusel2017gans} on COCO 2014~\cite{lin2014microsoft} validation dataset. We also plot the FID score of Stable Diffusion 1.5 and APDM.}
    \label{fig:naive_emp}
\end{figure}

\section{Additional Experiments}
\label{supp:additional_experiments}

\paragraph{Empirical Results about the limitation of Na\"ive Approach.}
In Section~\ref{supp:proof_derivation}, we theoretically demonstrated the fundamental limitations of na\"ive approach.
In the following part, we empirically validate those findings.
We applied the loss function of na\"ive approach, $\mathcal{L}_{adv}$ (Equation~\eqref{eq:L_adv}), to Stable Diffusion 1.5 with $N_{protect}=800$, as APDM.
We measured FID score every 100 iterations. 
As shown in Figure~\ref{fig:naive_emp}, as the optimization progresses,  the FID score consistently increases across all tested $\lambda$ values.
This degradation in quality occurs because the primary objective of $\mathcal{L}_{adv}$, minimizing $-\mathcal{L}_{simple}^{per}$ (\ie actively erasing related to the target for anti-personalization), becomes overly dominant.
Even though $\mathcal{L}_{ppl}$ is intended to preserve the generation performance, its effectiveness is clearly restricted by the optimized condition of $-\mathcal{L}_{simple}^{per}$.
This result aligns with our Theorem~\ref{theo:theorem1}, which suggests that the loss of each term in $\mathcal{L}_{adv}$ cannot be satisfied simultaneously.
Furthermore, when the weight $\lambda$ increases, one might expect a better preservation of the generative performance.
Although FID scores are relatively low with high $\lambda$ values (\eg $\lambda=10.0, 15.0$) in initial iterations, they still remain significantly high and can exhibit instability as training progresses. 
This suggests that our theorem is still valid in various $\lambda$.

\begin{table}[!t]
\caption{\textbf{Quantitative comparison on protection with image transformations.} We compared APDM with transformed images. For data poisoning baselines, we applied image transformation to perturbed images and we personalized Stable Diffusion on these transformed images. For APDM, we protected diffusion models on clean images and we conduct personalization on images that is transformed from clean images.}
\label{tab:transform}
\centering
\resizebox{\linewidth}{!}{%
\begin{tabular}{lcccgccg}
\toprule
\multirow{2}{*}{Methods} & \multirow{2}{*}{Transform.} & \multicolumn{3}{c}{DINO ($\downarrow$)} & \multicolumn{3}{c}{BRISQUE ($\uparrow$)} \\ \cmidrule{3-8} 
                         &     &\textit{``person''} &\textit{``dog''} & Avg. & \textit{``person''} & \textit{``dog''} & Avg. \\ \midrule
DreamBooth~\cite{ruiz2023dreambooth}& - & 0.6994 & 0.6056 & 0.6525 & 11.27 & 22.33 & 16.80 \\ \midrule
\multirow{3}{*}{AdvDM~\cite{liang2023adversarial}} & - & 0.5752 & 0.4247 & 0.4999 & 19.52 & 28.60 & 24.06 \\
                                & flip  & 0.5436 & 0.4538 & 0.4987 & 24.37 & 27.07 & 25.72 \\
                                 & blur & 0.6417 & 0.4524 & 0.5470 & 18.28 & 26.35 & 22.32 \\ \dashmidrule
\multirow{3}{*}{Anti-DreamBooth~\cite{van2023anti}} & - & 0.5254 & 0.4106 & 0.4680 & 26.90 & 30.23 & 28.56 \\
                                & flip  & 0.5976 & 0.4665 & 0.5321 & 26.76 & 29.19 & 27.97 \\
                                 & blur & 0.5487 & 0.4414 & 0.4951 & 24.37 & 28.91 & 26.64 \\ \dashmidrule
\multirow{3}{*}{SimAC~\cite{Wang2024simac}} & - & 0.4448 & 0.4374 & 0.4411 & 23.73 & 31.64 & 27.69 \\          
                                & flip  & 0.5083 & 0.4475 & 0.4779 & 26.56 & 29.46 & 28.01 \\
                                 & blur & 0.5323 & 0.4390 & 0.4856 & 20.40 & 31.27 & 25.83 \\ 
                                 \hdashline
\multirow{3}{*}{PAP~\cite{wan2024prompt}} & - & 0.6556 & 0.5120 & 0.5838 & 22.61 & 30.20 & 26.41 \\          
                                & flip  & 0.6564 & 0.5139 & 0.5852 & 22.51 & 27.81 & 25.16 \\
                                 & blur & 0.6708 & 0.5222 & 0.5965 & 24.37 & 27.83 & 26.10 \\ \dashmidrule
\multirow{3}{*}{\textbf{APDM (Ours)}}   & - & 0.1375 & 0.0959 & \textbf{0.1167} & 40.25 & 60.74 & \textbf{50.50} \\ 
                                & flip  & 0.1714  & 0.1194 & \textbf{0.1454} & 39.13 & 40.34 & \textbf{39.74} \\
                                 & blur &  0.1042 & 0.0823 & \textbf{0.0933} & 40.47 & 45.13 & \textbf{42.80} \\ \bottomrule
\end{tabular}%
}
\end{table}

\paragraph{Protection with Image Transformations.}
In Figure~\ref{fig:clean_all} and Table~\ref{tab:main} of the main paper, we compared APDM with baselines considering the existence and quantity of clean images.
Additionally, we also compared APDM with baselines using transformed images such as flipping and blurring.
Table~\ref{tab:transform} demonstrates that baselines fail to effectively protect personalization when transformations are applied to perturbed images.
In contrast, APDM exhibits robustness even under such image transformations.

\begin{figure}[!t]
    \centering
    \includegraphics[width=\linewidth]{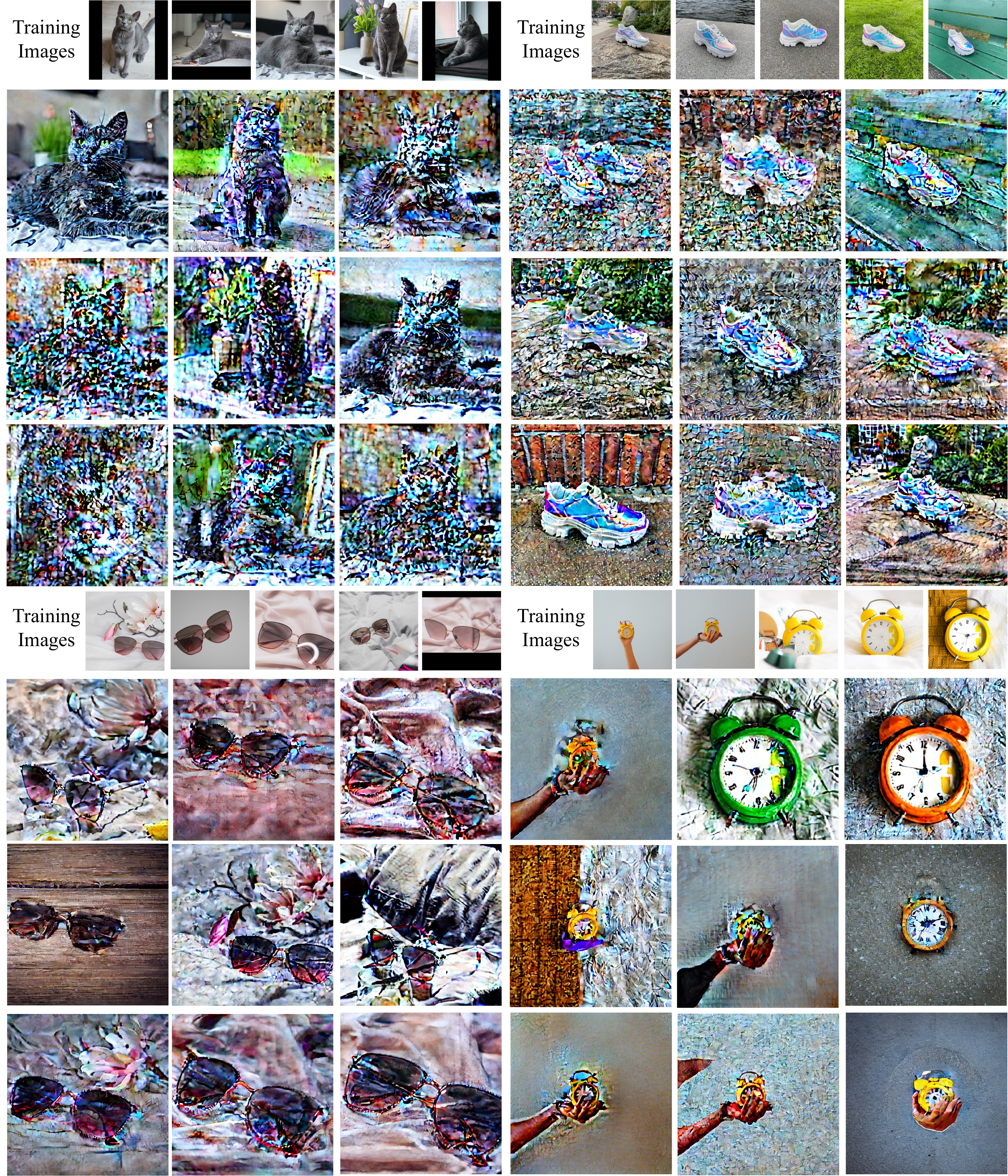}
    \caption{\textbf{Protection on other subjects.} We attempted to protect personalization on \textit{``cat''}, \textit{``sneaker''}, \textit{``glasses''}, and \textit{``clock''}.}
    \label{fig:protection_other_classes}
\end{figure}
\begin{table}[!t]
\caption{\textbf{Protection performance on other subjects.} In addition to experiments in the main paper, we evaluated APDM on different subjects. We tried to prevent personalization on \textit{``cat''}, \textit{``sneaker''}, \textit{``glasses''}, and \textit{``clock''}.}
\label{tab:protection_other_classes}
\centering
\resizebox{\linewidth}{!}{%
\begin{tabular}{lcccccccc}
\toprule
\multirow{2}{*}{Methods} &
  \multicolumn{4}{c}{DINO ($\downarrow$)} &
  \multicolumn{4}{c}{BRISQUE ($\uparrow$)} \\ \cmidrule{2-9} 
 &
  \textit{``cat''} &
  \textit{``sneaker''} &
  \textit{``glasses''} &
  \textit{``clock''}
   &
  \textit{``cat''} &
  \textit{``sneaker''} &
  \textit{``glasses''} &
  \textit{``clock''}
   \\ \midrule
DreamBooth~\cite{ruiz2023dreambooth}  & 0.4903 & 0.6110 & 0.6961 & 0.5359 & 25.32 & 23.14 & 19.01 & 13.82 \\ \dashmidrule
\textbf{APDM (Ours)} & \textbf{0.0414} & \textbf{0.2276} & \textbf{0.2893}& \textbf{0.1969} & \textbf{47.65} & \textbf{35.23} & \textbf{31.75} & \textbf{32.01} \\ \bottomrule
\end{tabular}%
}
\end{table}
\paragraph{Protection on Other Subjects.}
In the experiments presented in the main paper, we primarily considered two types of subjects: \textit{``person''} and \textit{``dog''}.
In Table \ref{tab:protection_other_classes} and Figure \ref{fig:protection_other_classes}, we explored the prevention of personalization on other subjects, such as \textit{``cat''}, \textit{``sneaker''}, \textit{``glasses''}, and \textit{``clock''}, demonstrating that APDM can be generally applied to protection of various subjects.

\begin{table}[!t]
\caption{\textbf{Protection performance of APDM on different personalization method, Custom Diffusion \cite{kumari2023multi}.} Unlike the experiments in the main paper, which used DreamBooth for personalization, we replaced the personalization method with Custom Diffusion. }
\label{tab:custom_diffusion}
\centering
\resizebox{0.6\linewidth}{!}{%
\begin{tabular}{lcccc}
\toprule
\multirow{2}{*}{Methods} & \multicolumn{2}{c}{DINO ($\downarrow$)}     & \multicolumn{2}{c}{BRISQUE ($\uparrow$)}    \\ \cmidrule{2-5} 
                 & \textit{``person''} & \textit{``dog''} & \textit{``person''} & \textit{``dog''} \\ \midrule
Custom Diffusion \cite{kumari2023multi} &  0.5320  & 0.5460  & 16.03 & 8.98 \\ \dashmidrule
\textbf{APDM (Ours)} & \textbf{0.2158} & \textbf{0.3202} & \textbf{34.61} & \textbf{33.09} \\ \bottomrule
\end{tabular}%
}
\end{table}

\section{Generalizability of APDM}
\label{supp:generalizablity}

\paragraph{Custom Diffusion.}
In our main paper, we mainly consider DreamBooth~\cite{ruiz2023dreambooth} as a personalization method.
Additionally, we utilized Custom Diffusion~\cite{kumari2023multi} as a variation of the personalization approach. 
In Table~\ref{tab:custom_diffusion}, we present the results of the Custom Diffusion experiments, and we conducted protection about \textit{``person''} and \textit{``dog''} similar to our main paper.
The results demonstrated that APDM can successfully prevent the personalization of Custom Diffusion, and show the applicability of APDM to other personalization methods.

\begin{table}[!t]
\caption{\textbf{Protection performance of APDM on different Stable Diffusion version, Stable Diffusion 2.1.
} In the experiments of the main paper, we primarily used Stable Diffusion 1.5. Additionally, we evaluated APDM based on different Stable Diffusion version.}
\label{tab:stable_diffusion_2_1}
\centering
\resizebox{0.6\linewidth}{!}{%
\begin{tabular}{lcccc}
\toprule
\multirow{2}{*}{Methods} & \multicolumn{2}{c}{DINO ($\downarrow$)} & \multicolumn{2}{c}{BRISQUE ($\uparrow$)} \\ \cmidrule{2-5} 
                         & \textit{``person''}   & \textit{``dog''}   & \textit{``person''}   & \textit{``dog''}   \\ \midrule
DreamBooth~\cite{ruiz2023dreambooth}  & 0.5773 & 0.5293 & 13.99 & 23.03 \\ \dashmidrule
\textbf{APDM (Ours)} & \textbf{0.2739} & \textbf{0.2178} & \textbf{39.72} & \textbf{42.69} \\ \bottomrule
\end{tabular}%
}
\end{table}
\paragraph{Stable Diffusion 2.1.}
APDM prevents personalization at the model level, and its applicability to different versions of the Stable Diffusion model is also important. 
In Table~\ref{tab:stable_diffusion_2_1}, we present experiments conducted on Stable Diffusion 2.1 to demonstrate the effectiveness of our approach on other diffusion models. 
We applied APDM to Stable Diffusion 2.1 and performed personalization with clean images using DreamBooth. 
The results indicate that APDM also performs robustly on Stable Diffusion 2.1, showing that our method is not restricted to a specific version of the diffusion model.

\begin{table}[!t]
\caption{\textbf{Protection performance on different unique identifier for personalization.} We conducted protection on \textit{``a photo of sks person''} or \textit{``a photo of sks dog''} and we tried to personalize diffusion models on \textit{``a photo of t@t person''} or \textit{``a photo of t@t dog''}.}
\label{tab:unique_identifier}
\centering
\resizebox{0.7\linewidth}{!}{%
\begin{tabular}{lccccc}
\toprule
\multirow{2}{*}{Methods} & \multirow{2}{*}{$[V^*]$} & \multicolumn{2}{c}{DINO ($\downarrow$)} & \multicolumn{2}{c}{BRISQUE ($\uparrow$)} \\ \cmidrule{3-6} 
            &                & \textit{``person''} & \textit{``dog''} & \textit{``person''} & \textit{``dog''} \\ \midrule
DreamBooth~\cite{ruiz2023dreambooth}  & ``t@t'' & 0.6774 & 0.4668 & 16.64 & 28.49 \\ \dashmidrule
\textbf{APDM (Ours)} & ``sks''$\rightarrow$``t@t'' & \textbf{0.3958} & \textbf{0.1981} & \textbf{29.90} & \textbf{40.69} \\ \bottomrule
\end{tabular}%
}
\end{table}
\paragraph{Prompt (Identifier) Mismatch.}
When an attacker performs personalization, they may use a different unique identifier (\eg \textit{``t@t''}) to capture the target subject. 
For example, during the protection process, we only show \textit{``a photo of sks person''}, while a different unique identifier may be used for personalization, such as \textit{``a photo of t@t person''}.
Similar to Van Le et al. \cite{van2023anti}, we also considered this prompt mismatch.
As shown in Table~\ref{tab:unique_identifier}, APDM can successfully protect against personalization attempts using \textit{``t@t''}. 
APDM successfully confuses the personalization process, preventing the identifier from capturing the target subject (\ie identity).

\begin{figure}[!t]
    \centering
    \includegraphics[width=0.9\linewidth]{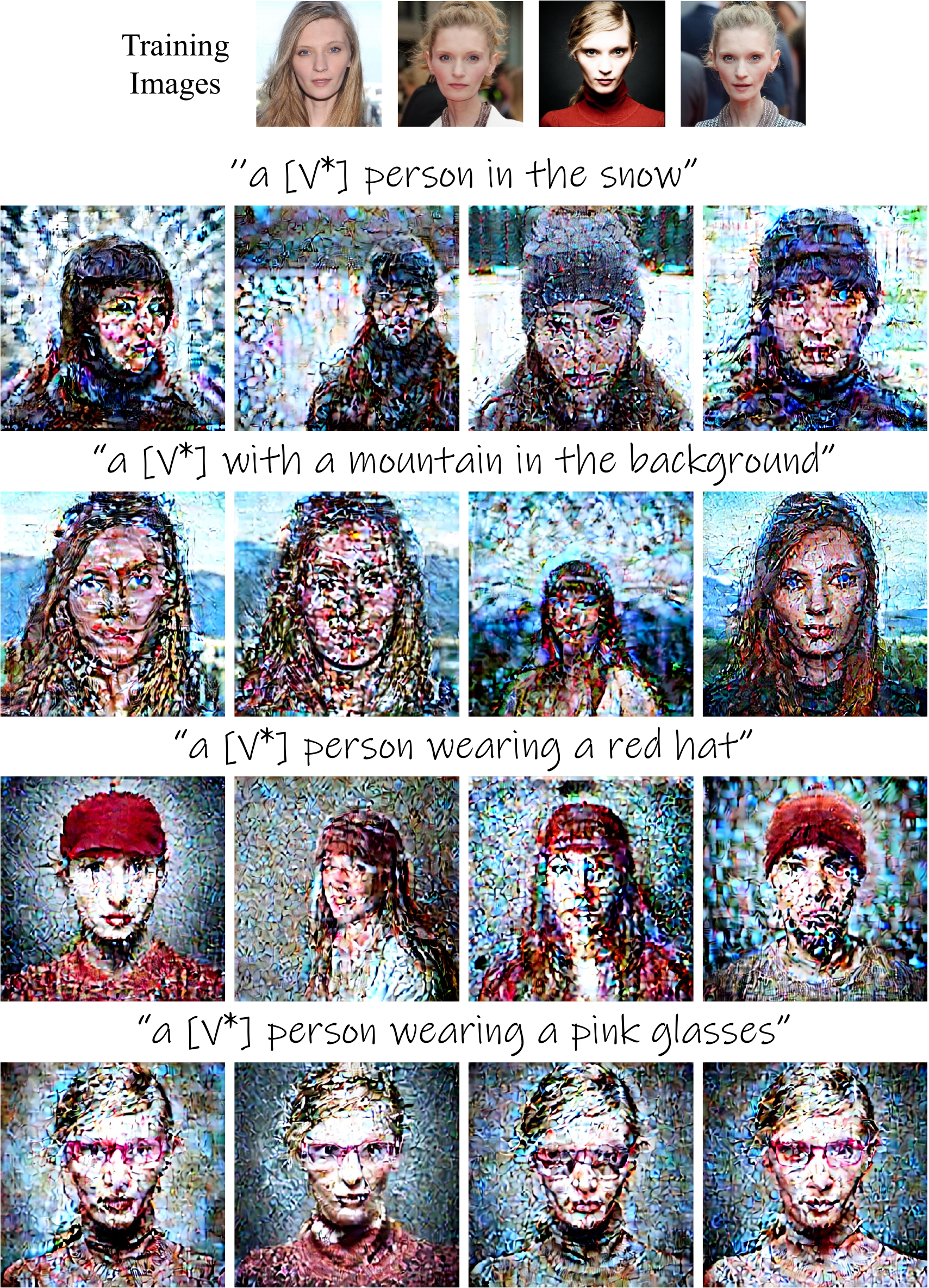}
    \caption{\textbf{Protection performance for diverse text prompts.} We visualized the generated outputs from diverse text prompts, such as \textit{``a [V*] person in the snow''} and \textit{``a [V*] person wearing a red hat''}.}
    \label{fig:diverse_prompts}
\end{figure}
\begin{table}[!t]
\caption{\textbf{Protection performance for diverse text prompts.} Unlike the experiments in the main paper, we evaluated APDM on diverse test prompts. Protection and personalization are conducted using \textit{``a photo of [V*] person''} or \textit{``a photo of [V*] dog''}, and we sampled images using the different set of text prompts.}
\label{tab:prompt_variation}
\centering
\resizebox{0.6\linewidth}{!}{%
\begin{tabular}{lccccc}
\toprule
\multirow{2}{*}{Methods} & \multicolumn{2}{c}{DINO ($\downarrow$)}     & \multicolumn{2}{c}{BRISQUE ($\uparrow$)} \\ \cmidrule{2-6} 
                         & \textit{``person''} & \textit{``dog''} & \textit{``person''}   & ``dog''   \\ \midrule
DreamBooth~\cite{ruiz2023dreambooth}  & 0.4081 & 0.4233 & 12.57 & 29.65 \\ \dashmidrule
\textbf{APDM (Ours)} & \textbf{0.1357} &  \textbf{0.1564} & \textbf{36.40} & \textbf{41.66} \\ \bottomrule
\end{tabular}%
}
\end{table}
\paragraph{Protection on Diverse Text Prompts.}
In the experiments presented in the main paper, we utilized simple text prompts for inference, such as \textit{``a photo of [V*] person''} and \textit{``a portrait of [V*] person.''}
In contrast to these experiments, we evaluated APDM using diverse prompts, such as \textit{``a photo of [V*] person in the jungle''} and \textit{``a [V*] person with a mountain background.''}
We adopted text prompts from the DreamBooth dataset \cite{ruiz2023dreambooth}.
Figure \ref{fig:diverse_prompts} and Table \ref{tab:prompt_variation} illustrate that APDM successfully prevents personalization, even under diverse prompt variations that differ from the text prompts used during the protection procedure.
This result highlights that APDM is even robust to diverse text prompt variation.
\vspace{-4mm}
\section{User Study}
\vspace{-1mm}
\label{supp:user_study}

\begin{figure}
    \centering
    \includegraphics[width=\linewidth]{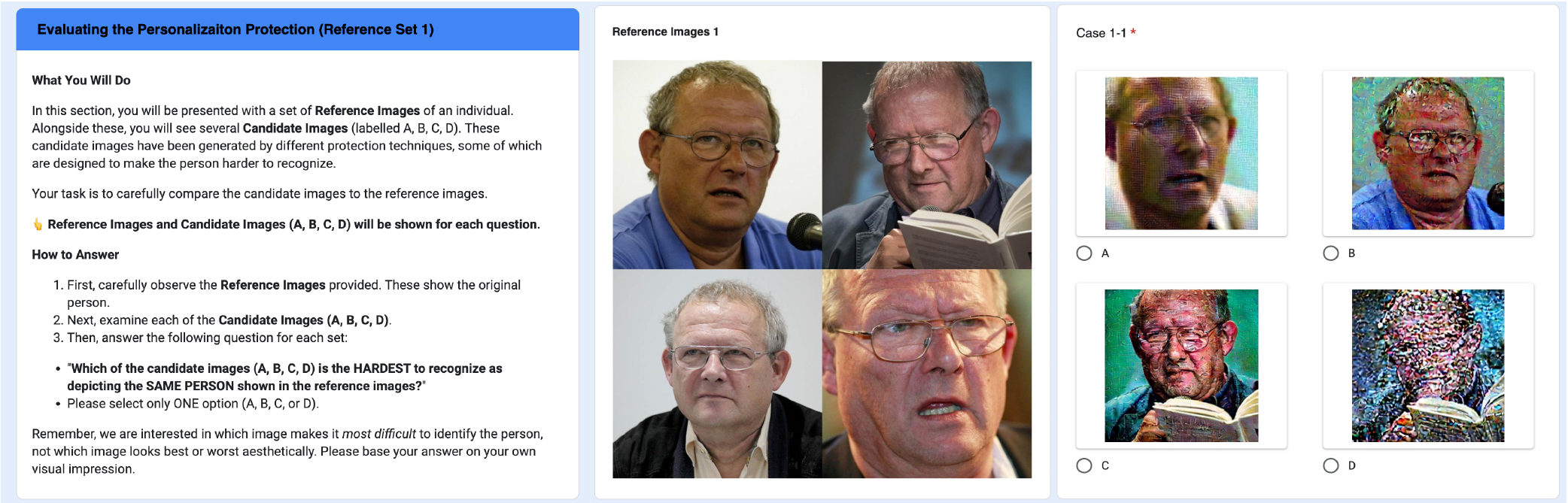}
    \caption{\textbf{A sample interface for our user study.} Left term is the descriptions of explanation about study. Middle term is a given reference images which used to capture the identity from participants. Right term is choices.}
    \vspace{-3mm}
    \label{fig:interface}
\end{figure}
\begin{table}[!t]
\caption{\textbf{Results of user study.} We count the percentage of votes for the comparisons and our method respectively. Every participants selected a sample that looks most different from the clean images.}
\label{tab:user_study}
\centering
\resizebox{0.8\linewidth}{!}{%
\begin{tabular}{ccccc}
\toprule
Methods     & Anti-DreamBooth~\cite{van2023anti}& SimAC~\cite{Wang2024simac}& PAP~\cite{wan2024prompt}& \textbf{APDM (Ours)} \\ \midrule
Protection & 7.08 \%           & 5.83 \%   & 1.04 \%   & \textbf{86.04 \%} \\
\bottomrule
\end{tabular}
}
\vspace{-2mm}
\end{table}

We conducted a user study to evaluate the preference of various protection methods in preventing subject recognition.
The specific questions and interface are illustrated in Figure~\ref{fig:interface}.
We presented four reference images for each subject to provide participants with clear identity information.
After viewing these, each participant chose an image based on the following question:
\begin{center}
    \textit{Which of the candidate images (A, B, C, D) is the \textbf{HARDEST} \\
    to recognize as depicting the \textbf{SAME PERSON} shown in the reference images?}
\end{center}

Candidate images were generated using a personalized diffusion model with different protection methods applied.
We utilized Stable Diffusion 1.5 and Dreambooth~\cite{ruiz2023dreambooth} as personalization method, which is the same as our experimental setting in main paper.
In this user study, we compared our proposed method, APDM, against Anti-Dreambooth~\cite{van2023anti}, SimAC~\cite{Wang2024simac}, and PAP~\cite{wan2024prompt}. 
For comparisons, we first generated perturbed images using each approach, and conducted personalization with these perturbed images. 
For APDM, we applied personalization using the model protected by our method.
After personalization, all images were generated using the prompt \textit{``a photo of [V*] person''}.
To ensure fairness, the same randomly sampled seed was used for generating all candidate images. 
The image sequence and the arrangement of choices are randomized to eliminate any bias. 

We collected responses from 25 voluntary adult participants regardless of gender. 
Participants were compensated \$0.125 USD per question, totaling \$2.50 USD, corresponding to hourly rate of \$7.26 USD.
On average, participants completed the study in about 20 minutes.
We did not collect any personal information from the participants.

As shown in Table~\ref{tab:user_study}, APDM achieved significantly higher user preference (\ie was selected more often as the hardest to recognize) than other comparisons. 
These results indicate that our method not only addresses limitations of data-centric approaches but also achieves a substantial improvement in protection performance. 
\section{Additional Qualitative Results}
\label{supp:additional_qualitative}
\begin{figure}[!t]
    \centering
    \includegraphics[width=\linewidth]{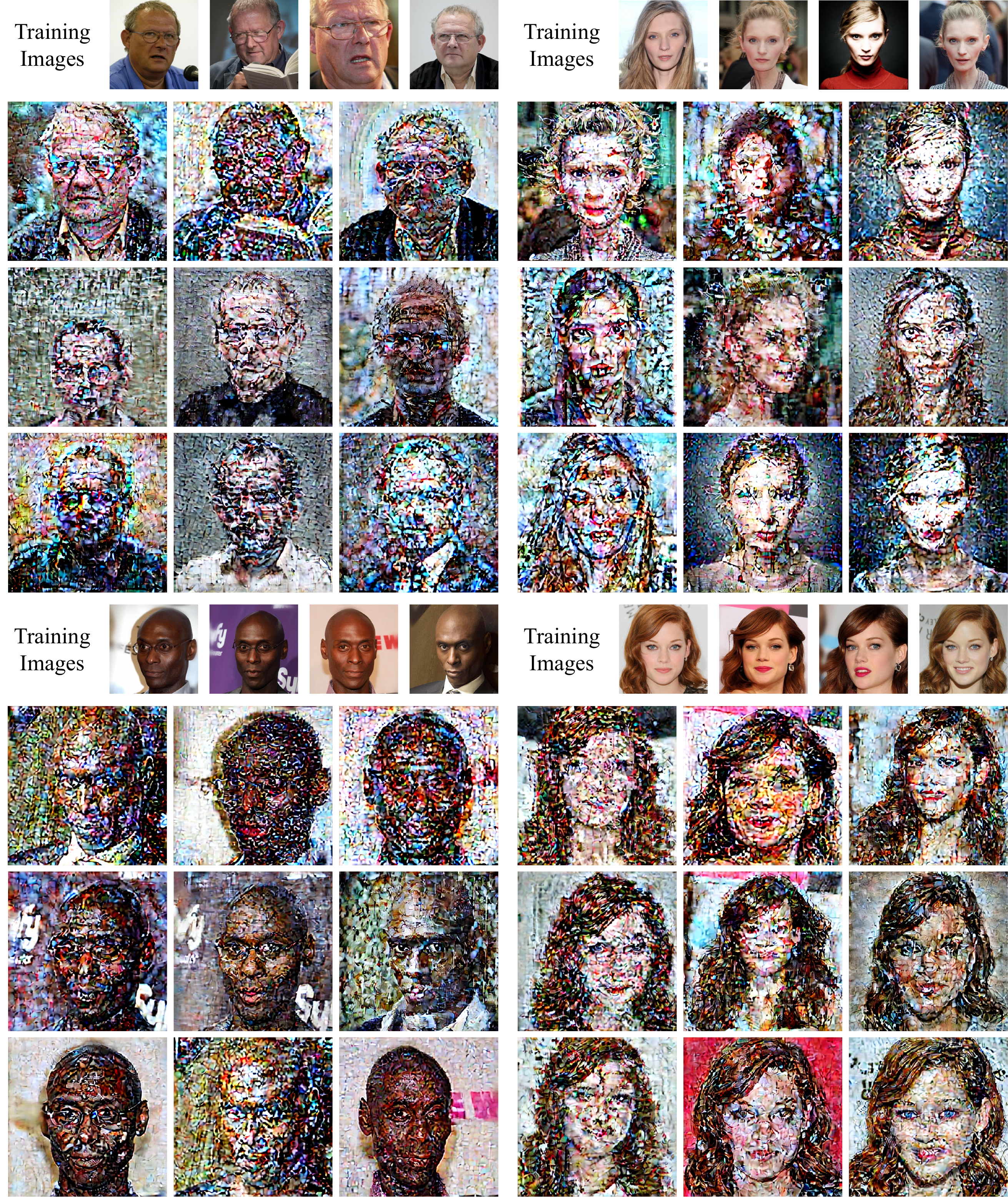}
    \caption{\textbf{Additional Qualitative Results on Protection (\textit{``person''}).}}
    \label{fig:additional_qual_person}
\end{figure}

\begin{figure}[!t]
    \centering
    \includegraphics[width=\linewidth]{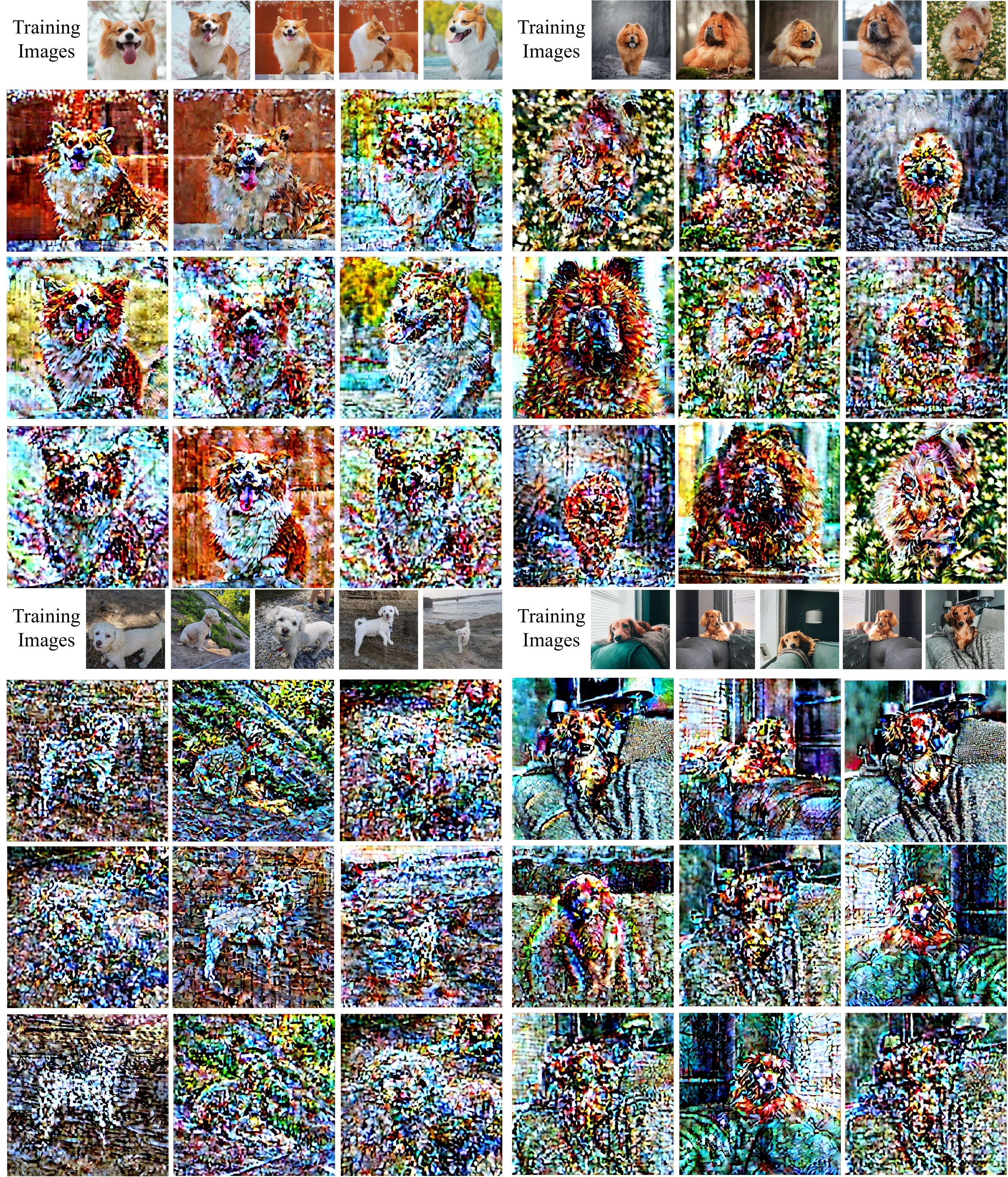}
    \caption{\textbf{Additional Qualitative Results on Protection (\textit{``dog''}).}}
    \label{fig:additional_qual_dog}
\end{figure}

\paragraph{Additional Protection Results.}
In Figure~\ref{fig:clean_all} and Table~\ref{tab:main} of the main paper, we conducted quantitative and qualitative experiments, respectively.
We attached additional qualitative results in Figure \ref{fig:additional_qual_person} and Figure \ref{fig:additional_qual_dog}, including protection results on various subjects of person and dog.
The experimental results highlight again that APDM can effectively protect personalization against diverse subjects, producing images of a lot of artifacts or containing different instances.
\clearpage
\clearpage
\section{Additional Explanation of Motivation}
\label{supp:add_motivate}
\vspace{-2mm}
Figure~\ref{fig:motivate} in our main paper presents the motivation for our work and briefly describes key issues, including the impractical assumptions of existing approaches, easy circumvention, user burdens, and conflict with regularization.
This section provides a more detailed explanation of these limitations to facilitate clearer understanding.

We first criticize the impracticality of the existing literature.
In daily life, individuals frequently take pictures or are photographed.
For example, they often take selfies for social media or capture images of their identification documents.
Such images, which we refer to as \textit{``User's Photos''} (as depicted in Figure~\ref{fig:motivate} of our main paper), are those that users are consciously aware of and possess. 
Consequently, users have the opportunity to apply protection methods (\ie data poisoning approach) to these photos if they want.
In contrast, \textit{``Unintended Capture''} refers to images of individuals taken without their explicit recognition or control over their subsequent use.
This scenario presents a critical vulnerability, as these unintentionally captured images can be exploited as unprotected, \textit{``clean''} data by malicious users.

As shown in Table~\ref{tab:main} of our main paper, the presence of clean (unprotected) images can significantly degrade the effectiveness of data poisoning techniques, allowing for easy bypass of protection.
Furthermore, even when images are perturbed (\ie poisoned), their protective effect is vulnerable to various common image transformations that frequently occur in real-world scenarios (as also shown in Table~\ref{tab:transform}).
These transformations can weaken or negate the intended poisoning effect.
These limitations reveal that, without strong (and often impractical) assumptions about the unavailability of clean images or the absence of transformations, existing protection methods exhibit restricted performance.

Regarding the user burden associated with implementing such techniques, most individuals are unfamiliar with implementation of AI technique.
Establishing appropriate hardware environments (\eg GPU servers) and configuring complex software environments (\eg managing numerous libraries and their dependencies) present a significant initial hurdle.
Even if these challenges are overcome, non-expert users still face substantial obstacles in utilizing protection methods.
These include a lack of fundamental understanding of the protection mechanisms themselves, insufficient understanding in necessary programming languages (such as Python), and inadequate debugging skills to troubleshoot issues.
These technical components are crucial for successful implementation of protection methods, yet their complexity also acts as a significant barrier, preventing widespread adoption by the general public.

The user-centric nature of existing data poisoning methods inherently conflicts with privacy regulations such as the General Data Protection Regulation (GDPR)~\cite{voigt2017gdpr}. 
The GDPR places the duty for privacy protection on service providers (\ie model owners) to ensure a user's request.
However, data poisoning approaches are ill-suited for service providers to fulfill this responsibility.
These methods typically operate at the individual image level, requiring modifications to user data before they interact with the model. 
Service providers, in contrast, primarily manage the model itself.
This operational disparity highlights why such user-side defenses are impractical for providers, underscoring the critical need for alternative approaches. 
To alleviate this, we propose a novel framework APDM, which empowers service providers to effectively manage and enforce anti-personalization directly within their systems, aligning with their responsibilities under privacy regulations and enabling a more scalable and reliable means of privacy protection.

\vspace{-3mm}
\section{Limitation and Broader Impacts}
\vspace{-2mm}
\label{supp:limitation}
In this work, we focused on protecting the personalization of a specific subject at the model level.
APDM offers a significant step towards more robust and practical privacy protection in personalization of diffusion model. 
By enabling direct, model-level anti-personalization, it empowers service providers to better comply with privacy regulations and reduces the burden on individual users to protect their own data. 
This could foster greater trust and safer use of powerful generative models in various applications.

While APDM effectively safeguards personalization for a single subject, real-world scenarios often require the protection of multiple subjects simultaneously.
Additionally, there may be a need to incorporate protection for new subjects into models that are already safeguarded.
Addressing these challenges presents an opportunity for future research, including multi-concept personalization protection and continual personalization safeguarding.


\end{document}